\documentclass{article} 
\usepackage{nips12submit_e,times}
\usepackage{times}
\usepackage{array}
\usepackage[ruled,vlined]{algorithm2e}
\usepackage{amsmath}
\usepackage{amssymb}
\usepackage{graphicx}
\usepackage{caption}
\usepackage{subcaption}
\usepackage{wrapfig}
\usepackage[named]{algo}
\usepackage{slashbox}
\usepackage{helvet}
\usepackage{courier}
\usepackage{epsfig}
\usepackage{amsfonts}
\usepackage{amsthm}

\newcommand{\argmax}{ \operatorname*{arg\,max}}
\newtheorem{theorem}{Theorem}

\newtheorem{definition}{Definition}

\newcommand{\squishlisttwo}{
 \begin{list}{$\bullet$}
  { \setlength{\itemsep}{3pt}
     \setlength{\parsep}{0pt}
    \setlength{\topsep}{0pt}
    \setlength{\partopsep}{0pt}
    \setlength{\leftmargin}{1em}
    \setlength{\labelwidth}{1.5em}
    \setlength{\labelsep}{0.5em} } }

\newcommand{\squishend}{
  \end{list}  }

\title{Towards Practical Planning to Predict and Exploit Intentions for Interacting with Self-Interested Agents}

\author{
Trong Nghia Hoang \& Kian Hsiang Low\\
Department of Computer Science, National University of Singapore\\
Republic of Singapore\\
\{nghiaht, lowkh\}@comp.nus.edu.sg}

\nipsfinalcopy 

\begin{document}

\newcounter{sol} 
\setcounter{sol}{1} 

\maketitle
\begin{abstract}
A key challenge in non-cooperative multi-agent systems is that of developing efficient planning algorithms for intelligent agents to interact and perform effectively among boundedly rational, self-interested agents (e.g., humans). The practicality of existing works addressing this challenge is being undermined due to either the restrictive assumptions of the other agents' behavior, the failure in accounting for their rationality, or the prohibitively expensive cost of modeling and predicting their intentions. To boost the practicality of research in this field, we investigate how intention prediction can be efficiently exploited and made practical in planning, thereby leading to efficient intention-aware planning frameworks capable of predicting the intentions of other agents and acting optimally with respect to their predicted intentions. We show that the performance losses incurred by the resulting planning policies are linearly bounded by the error of intention prediction. Empirical evaluations through a series of stochastic games demonstrate that our policies can achieve better and more robust performance than the state-of-the-art algorithms.
\end{abstract}

\section{Introduction}
\label{sect:intro}
A fundamental challenge in non-cooperative \emph{multi-agent systems} (MAS) is that of designing intelligent agents that can efficiently plan their actions under uncertainty to interact and perform effectively among boundedly rational\footnote{Boundedly rational agents are subject to limited cognition and time in making decisions \cite{Reinhard2002}.}, self-interested agents (e.g., humans). Such a challenge is posed by many real-world applications \cite{Hansen2004}, which include automated electronic trading markets where software agents interact, and traffic intersections where autonomous cars have to negotiate with human-driven vehicles to cross them, among others. These applications can be modeled as partially observable stochastic games (POSGs) in which the agents are self-interested (i.e., non-cooperative) and do not necessarily share the same goal, thus invalidating the use of planning algorithms developed for coordinating cooperative agents (i.e., solving POSGs with common payoffs) \cite{Nair2003,Seuken2007b,Spaan2011}. Existing planning frameworks for non-cooperative MAS can be generally classified into:\vspace{1mm}

\noindent
{\bf Game-theoretic frameworks.} Based on the well-founded classical game theory, these multi-agent planning frameworks \cite{Hansen2004,Hu1998}\footnote{The learning framework of Hu and Wellman [1998] trivially reduces to planning when the transition model is known a priori.} characterize the agents' interactions in a POSG using solution concepts such as Nash equilibrium. 
Such frameworks suffer from the following drawbacks: (a) Multiple equilibria may exist, (b) only the optimal actions corresponding to the equilibria are specified, and (c) they assume that the agents do not collaborate to beneficially deviate from the equilibrium (i.e., no coalition), which is often violated by human agents.\vspace{1mm}

\noindent
{\bf Decision-theoretic frameworks.} Unafflicted by the drawbacks of game-theoretic approaches, they extend single-agent decision-theoretic planning frameworks such as the \emph{Markov decision process} (MDP) and \emph{partially observable Markov decision process} (POMDP) to further characterize interactions with the other self-interested agents in a POSG.
In particular, the \emph{interactive POMDP} (I-POMDP) framework \cite{Doshi2009,Doshi2008,Doshi2005,Doshi2006} is proposed to explicitly account for the bounded rationality of self-interested agents: It replaces POMDP's flat beliefs over the physical states with interactive beliefs over both the physical states and the other agent's beliefs. Empowered by such an enriched, highly expressive belief space, I-POMDP can explicitly model and predict the other agent's intention (i.e., mixed strategy) under partial observability.\vspace{1mm}

However, solving I-POMDP is prohibitively expensive due to the following computational difficulties \cite{Doshi2008,Doshi2005}: (a) {\bf Curse of dimensionality} -- since I-POMDP's interactive belief is over the joint space of physical states and the other agent's beliefs (termed interactive state space in \cite{Doshi2008}), its dimension can be extremely large and possibly infinite; (b) {\bf curse of history} -- similar to POMDP, I-POMDP's policy space grows exponentially with the length of planning horizon; and (c) {\bf curse of nested reasoning} -- 
as I-POMDP utilizes a nested structure within our agent's belief space to represent its belief over the other agent's belief and the other agent's belief over our agent's belief and so on, it aggravates the effects of the other two curses \cite{Doshi2008}.

To date, a number of approximate I-POMDP techniques \cite{Doshi2009,Doshi2008,Doshi2006} have been proposed to mitigate some of the above difficulties. Notably, \emph{Interactive Particle Filtering} (I-PF) \cite{Doshi2009} focused on alleviating the curse of dimensionality by generalizing the particle filtering technique to accommodate the multi-agent setting while \emph{Interactive Point-based Value Iteration} \cite{Doshi2008} (I-PBVI) aimed at relieving the curse of history by generalizing the well-known point-based value iteration (PBVI) \cite{Pineau2003} to operate in the interactive belief space. 
Unfortunately, I-PF fails to address the curse of history and it is not clear how PBVI or other sampling-based algorithm can be modified to work with a particle representation of interactive beliefs, whereas I-PBVI suffers from the curse of dimensionality because its dimension of interactive belief grows exponentially with the length of planning horizon of the other agent (Section~\ref{sect:iap}).
Using interactive beliefs, it is therefore not known whether it is even possible to jointly lift both curses, for example, by extending I-PF or I-PBVI.
Furthermore, they do not explicitly account for the curse of nested reasoning. 
As a result, their use has been restricted to small, simple problems \cite{Brenda2010} (e.g., multiagent Tiger \cite{Doshi2005,Nair2003,Doshi2008}).

To tractably solve larger problems, existing approximate I-POMDP techniques such as I-PF and I-PBVI have to significantly reduce the quality of approximation and impose restrictive assumptions (Section~\ref{sect:iap}), or risk not producing a policy at all with the available memory of modern-day computers. 
This naturally raises the concern of whether the resulting policy can still perform well or not under different partially observable environments, as investigated in Section~\ref{sect:exp2}. 
Since such drastic compromises in solution quality are necessary of approximate I-POMDP techniques to tackle a larger problem directly, it may be worthwhile to instead consider formulating an approximate version of the problem with a less sophisticated structural representation such that it allows an exact or near-optimal solution policy to be more efficiently derived. 
More importantly, can the induced policy perform robustly against errors in modeling and predicting the other agent's intention? If we are able to formulate such an approximate problem, 
the resulting policy can potentially perform better than an approximate I-POMDP policy in the original problem while incurring significantly less planning time.

Our work in this paper investigates such an alternative: 
We first develop a novel intention-aware \emph{nested MDP} framework (Section~\ref{sect:ip}) for planning in fully observable multi-agent environments. Inspired by the cognitive hierarchy model of games \cite{Camerer2004}, nested MDP constitutes a recursive reasoning formalism to predict the other agent's intention and then exploit it to plan our agent's optimal interaction policy. 
Its formalism is by no means a reduction of I-POMDP. We show that nested MDP incurs linear time in the planning horizon length and reasoning depth. Then, we propose an I-POMDP Lite framework (Section~\ref{sect:iap}) for planning in partially observable multi-agent environments that, in particular, exploits a practical structural assumption: The intention of the other agent is driven by nested MDP, which is demonstrated theoretically to be an effective surrogate of its true intention when the agents have fine sensing and actuation capabilities. 
This assumption allows the other agent's intention to be predicted efficiently and, consequently, I-POMDP Lite to be solved efficiently in polynomial time, hence lifting the three curses of I-POMDP.
As demonstrated empirically, it also improves I-POMDP Lite's robustness in planning performance by overestimating the true sensing capability of the other agent. We provide theoretical performance guarantees of the nested MDP and I-POMDP Lite policies that improve with decreasing error of intention prediction (Section~\ref{sect:thr}). We extensively evaluate our frameworks through experiments involving a series of POSGs that have to be modeled using a significantly larger state space (Section~\ref{sect:exp2}).

\section{Nested MDP}
\label{sect:ip}
Given that the environment is fully observable, our proposed nested MDP framework can be used to predict the other agent's strategy and such predictive information is then exploited to plan our agent's optimal interaction policy. Inspired by the cognitive hierarchy model of games \cite{Camerer2004}, it constitutes a well-defined recursive reasoning process that comprises $k$ levels of reasoning. At level $0$ of reasoning, our agent simply believes that the other agent chooses actions randomly and computes its best response by solving a conventional MDP that implicitly represents the other agent's actions as stochastic noise in its transition model. At higher reasoning levels $k \geq 1$, our agent plans its optimal strategy by assuming that the other agent's strategy is based only on lower levels $0, 1, \ldots, k - 1$ of reasoning. In this section, we will formalize nested MDP and show that our agent's optimal policy at level $k$ can be computed recursively.
\vspace{1.5mm}

\noindent
{\bf Nested MDP Formulation.}
Formally, nested MDP for agent $t$ at level $k$ of reasoning is defined as a tuple $M_{t}^{k} \triangleq \left(S, U, V, T, R, \{\pi^{i}_{\mbox{-}t}\}_{i=0}^{k-1}, \phi\right)$ where
$S$ is a set of all possible states of the environment;
$U$ and $V$ are, respectively, sets of all possible actions available to agents $t$ and $\mbox{-}t$;
$T: S \times U \times V \times S \rightarrow [0, 1]$ denotes the probability $Pr(s'|s,u,v)$ of going from state $s\in S$ to state $s'\in S$ using agent $t$'s action $u\in U$ and agent $\mbox{-}t$'s action $v\in V$;
$R: S \times U \times V \rightarrow \mathbb{R}$ is a reward function of agent $t$;
$\pi^{i}_{\mbox{-}t} : S \times V \rightarrow [0, 1]$ is a reasoning model of agent $\mbox{-}t$ at level $i < k$, as defined later in (\ref{eq:m3});
and $\phi \in (0, 1) $ is a discount factor.
\vspace{1.5mm}

\noindent
{\bf Nested MDP Planning.}
The optimal $(h+1)$-step-to-go value function of nested MDP $M_{t}^{k}$ at level $k \geq 0$ for agent $t$ satisfies the following Bellman equation:\vspace{-1mm}
\begin{equation}
\hspace{-1.8mm}
\begin{array}{rl}
U^{k, h + 1}_t(s) \hspace{-0mm}\triangleq &\hspace{-2mm} \displaystyle\max_{u \in U}\sum_{v \in V}\widehat{\pi}^{k}_{\mbox{-}t} (s, v)\ Q^{k, h+1}_t(s, u, v)\\
Q^{k, h+1}_t(s, u, v) \hspace{-0mm}\triangleq &\hspace{-2mm} \displaystyle R(s, u,v) + \phi\sum_{s' \in S}T(s, u,v, s')\ U^{k, h}_t(s')\vspace{-3mm}
\end{array}
\label{eq:m1}
\end{equation}
where the mixed strategy $\widehat{\pi}^{k}_{\mbox{-}t}$ of the other agent $\mbox{-}t$ for $k>0$ is predicted as
\begin{equation}
\widehat{\pi}^{k}_{\mbox{-}t}(s, v) \triangleq \left\{
\begin{array}{cl}
\sum_{i=0}^{k-1}p(i)\pi^{i}_{\mbox{-}t}(s, v) & \text{if $k > 0$},\vspace{1mm}\\
|V|^{-1} & \text{otherwise}.
\end{array} \right. 
\label{eq:m2}
\end{equation}
where the probability $p(i)$ (i.e., $\sum_{i=0}^{k-1} p(i) = 1$) specifies how likely agent $\mbox{-}t$ will reason at level $i$; a uniform distribution is assumed when there is no such prior knowledge. Alternatively, one possible direction for future work is to learn $p(i)$ using multi-agent reinforcement learning techniques such as those described in \cite{Chalkiadakis2003,NghiaIJCAI13b}. 
At level $0$, agent $\mbox{-}t$'s reasoning model $\pi^{0}_{\mbox{-}t}$ is induced by solving $M_{\mbox{-}t}^0$. 
To obtain agent $\mbox{-}t$'s reasoning models $\{\pi^{i}_{\mbox{-}t}\}_{i=1}^{k-1}$ at levels $i=1,\ldots,k-1$, let $Opt_{\mbox{-}t}^{i}(s)$ be the set of agent $\mbox{-}t$'s optimal actions for state $s$ induced by solving its nested MDP $M_{\mbox{-}t}^{i}$, which recursively involves building agent $t$'s reasoning models $\{\pi^{l}_{t}\}_{l=0}^{i-1}$ at levels $l=0, 1, \ldots, i-1$, by definition. Then, 
\begin{equation}
\pi^{i}_{\mbox{-}t}(s, v) \triangleq \left\{
\begin{array}{cl}
\displaystyle |Opt_{\mbox{-}t}^{i}(s)|^{-1} & \text{if $v \in Opt_{\mbox{-}t}^{i}(s)$},\vspace{1mm}\\
0 & \text{otherwise}.
\end{array} \right.
\label{eq:m3}
\end{equation}
After predicting agent $\mbox{-}t$'s mixed strategy $\widehat{\pi}^{k}_{\mbox{-}t}$ (\ref{eq:m2}), agent $t$'s optimal policy (i.e., reasoning model) $\pi^{k}_{t}$ at level $k$ can be induced by solving its corresponding nested MDP $M_{t}^{k}$ (\ref{eq:m1}).
\vspace{1.5mm}

\noindent
{\bf Time Complexity.} Solving $M_{t}^{k}$ involves solving $\{M_{\mbox{-}t}^{i}\}_{i=0}^{k - 1}$, which, in turn, requires solving $\{M_{t}^{i}\}_{i=0}^{k-2}$, and so on. Thus, solving $M_{t}^{k}$ requires solving $M_{t}^{i}$ ($i =0,\ldots, k - 2$) and $M_{\mbox{-}t}^{i}$ ($i=0,\ldots, k - 1$), that is, $\mathcal{O}\hspace{-0.8mm}\left(k\right)$ nested MDPs. Given $\widehat{\pi}^{k}_{\mbox{-}t}$, the cost of deriving agent $t$'s optimal policy grows linearly with the horizon length $h$ as the backup operation (\ref{eq:m1}) has to be performed $h$ times. In turn, each backup operation incurs $\mathcal{O}\hspace{-0.8mm}\left(|S|^2\right)$ time given that $|U|$ and $|V|$ are constants. Then, given agent $\mbox{-}t$'s profile of reasoning models $\{\pi^{i}_{\mbox{-}t}\}_{i=0}^{k-1}$, predicting its mixed strategy $\widehat{\pi}^{k}_{\mbox{-}t} (s, v)$ (\ref{eq:m2}) incurs $\mathcal{O}\hspace{-0.8mm}\left(k\right)$ time. Therefore, solving agent $t$'s nested MDP $M_{t}^{k}$ (\ref{eq:m1}) or inducing its corresponding reasoning model $\pi^{k}_{t}$ incurs $\mathcal{O}\hspace{-0.8mm}\left(kh|S|^2\right)$.
\section{Intention-Aware POMDP}
\label{sect:iap}
To tackle partial observability, it seems obvious to first consider generalizing the recursive reasoning formalism of nested MDP. This approach yields two practical complications: (a) our agent's belief over both the physical states and the other agent's beliefs (i.e., a probability distribution over probability distributions) has to be modeled, and (b) the other agent's mixed strategy has to be predicted for each of its infinitely many possible beliefs. Existing approximate I-POMDP techniques address these respective difficulties by (a) using a finite particle representation like I-PF \cite{Doshi2009} or (b) constraining the interactive state space $IS$ to $IS' = S \times \mbox{Reach}(B, h)$ like I-PBVI \cite{Doshi2008} where $\mbox{Reach}(B, h)$ includes the other agent's beliefs reachable from a finite set $B$ of its candidate initial beliefs over horizon length $h$.

However, recall from Section~\ref{sect:intro} that since I-PF suffers from the curse of history, the particle approximation of interactive beliefs has to be made significantly coarse to solve larger problems tractably, thus degrading its planning performance. I-PBVI, on the other hand, is plagued by the curse of dimensionality due to the need of constructing the set $\mbox{Reach}(B, h)$ whose size grows exponentially with $h$. As a result, it cannot tractably plan beyond a few look-ahead steps for even the small test problems in Section~\ref{sect:exp2}. Furthermore, it imposes a restrictive assumption that the true initial belief of the other agent, which is often not known in practice, needs to be included in $B$ to satisfy the absolute continuity condition of interactive beliefs \cite{Doshi2008} (see Appendix C for more details). So, I-PBVI may not perform well under practical environmental settings where a long planning horizon is desirable or the other agent's initial belief is not included in $B$. For I-PF and I-PBVI, the curse of nested reasoning aggravates the effects of other curses.

Since predicting the other agent's intention using approximate I-POMDP techniques is prohibitively expensive, it is practical to consider a computationally cheaper yet credible information source providing its intention such as its nested MDP policy. Intuitively, such a policy describes the intention of the other agent with full observability who believes that our agent has full observability as well. Knowing the other agent's nested MDP policy is especially useful when the agents' sensing and actuation capabilities are expected to be good (i.e., accurate observation and transition models), as demonstrated in the following simple result:
\begin{theorem}
Let $\widehat{Q}_{\mbox{-}t}^{n}(s,v)\triangleq |U|^{-1}\sum_{u\in U}Q_{\mbox{-}t}^{0,n}(s,v,u)$ and $\widehat{Q}_{\mbox{-}t}^{n}(b,v)$ denote $n$-step-to-go values of selecting action $v\in V$ in state $s\in S$ and belief $b$, respectively, for the other agent $\mbox{-}t$ using nested MDP and I-POMDP at reasoning level $0$ (i.e., MDP and POMDP). 
If $b(s) \geq 1 - \epsilon$ and\vspace{-2mm} 
$$\displaystyle\forall \left(s, v, u\right) \exists \left(s', o\right) Pr(s' | s, v, u) \hspace{-1mm}\geq\hspace{-1mm} 1 - \frac{\epsilon}{2}\ \wedge\ Pr(o|s,v) \hspace{-1mm}\geq\hspace{-1mm} 1 - \frac{\epsilon}{2}\vspace{-2mm}$$ 
for some  $\epsilon\geq 0$, then
\begin{equation}
\left|\widehat{Q}_{\mbox{-}t}^{n}(s,v) - \widehat{Q}_{\mbox{-}t}^{n}(b,v)\right| \leq \epsilon\ \mathcal{O}\hspace{-0.8mm}\left(\frac{R_{\max} - R_{\min}}{1 - \phi}\right)\vspace{-0.3mm}\label{eq:0}
\end{equation}
where $R_{\max}$ and $R_{\min}$ denote agent $\mbox{-}t$'s maximum and minimum immediate payoffs, respectively.
\label{thm:0}
\end{theorem}
Its proof is given in Appendix A. Following from Theorem~\ref{thm:0}, we conjecture that, as $\epsilon$ decreases (i.e., observation and transition models become more accurate), the nested MDP policy ${\pi}^{0}_{\mbox{-}t}$ of the other agent is more likely to approximate the exact I-POMDP policy closely.
Hence, the nested MDP policy serves as an effective surrogate of the exact I-POMDP policy (i.e., true intention) of the other agent if the agents have fine sensing and actuation capabilities; such a condition often holds for typical real-world environments.

Motivated by the above conjecture and Theorem~\ref{thm:0}, we propose an alternative I-POMDP Lite framework by exploiting the following structural assumption: 
The intention of the other agent is driven by nested MDP.
This assumption allows the other agent's intention to be predicted efficiently by computing its nested MDP policy, thus lifting I-POMDP's curse of nested reasoning (Section~\ref{sect:ip}). More importantly, it enables both the curses of dimensionality and history to be lifted, which makes solving I-POMDP Lite very efficient, as explained below.
Compared to existing game-theoretic frameworks \cite{Hu1998,Littman1994} which make strong assumptions of the other agent's behavior, our assumption is clearly less restrictive. Unlike the approximate I-POMDP techniques, it does not cause I-POMDP Lite to be subject to coarse approximation when solving larger problems, which can potentially result in better planning performance. Furthermore, by modeling and predicting the other agent's intention using nested MDP, I-POMDP Lite tends to overestimate its true sensing capability and can therefore achieve a more robust performance than I-PBVI using significantly less planning time under different partially observable environments (Section~\ref{sect:exp2}).
\vspace{1.5mm}

\noindent
{\bf I-POMDP Lite Formulation.} Our I-POMDP Lite framework constitutes an integration of the nested MDP for predicting the other agent's mixed strategy into a POMDP for tracking our agent's belief in partially observable environments. 
Naively, this can be achieved by extending the belief space to $\Delta(S \times V)$ (i.e., each belief $b$ is now a probability distribution over the state-action space $S \times V$) and solving the resulting augmented POMDP. The size of representing each belief therefore becomes $\mathcal{O}(|S||V|)$ 
(instead of $\mathcal{O}(|S|)$), which consequently increases the cost of processing each belief (i.e., belief update). Fortunately, our I-POMDP Lite framework can alleviate this extra cost: By factorizing $b(s, v) = b(s)\ \widehat{\pi}^k_{\mbox{-}t}(s, v)$, the belief space over $S \times V$ can be reduced to one over $S$ because the predictive probabilities $\widehat{\pi}^k_{\mbox{-}t}(s, v)$ (\ref{eq:m2}) (i.e., predicted mixed strategy of the other agent) are derived separately in advance by solving nested MDPs. 
This consequently alleviates the curse of dimensionality pertaining to the use of interactive beliefs, as discussed in Section~\ref{sect:intro}. 
Furthermore, such a reduction of the belief space decreases the time and space complexities and typically allows an optimal policy to be derived faster in practice: the space required to store $n$ sampled beliefs is only $\mathcal{O}(n|S| + |S||V|)$ instead of $\mathcal{O}(n|S||V|)$.

Formally, I-POMDP Lite (for our agent $t$) is defined as a tuple $(S, U, V, O, T, Z, R, \widehat{\pi}^{k}_{\mbox{-}t}, \phi, b_0)$ where
$S$ is a set of all possible states of the environment;
$U$ and $V$ are sets of all actions available to our agent $t$ and the other agent $\mbox{-}t$, respectively;
$O$ is a set of all possible observations of our agent $t$;
$T: S \times U \times V \times S \rightarrow [0, 1]$ is a transition function that depends on the agents' joint actions;
$Z: S \times U \times O \rightarrow [0, 1]$ denotes the probability $Pr(o|s',u)$ of making observation $o\in O$ in state $s'\in S$ using our agent $t$'s action $u\in U$;
$R: S \times U \times V \rightarrow \mathbb{R}$ is the reward function of agent $t$;
$\widehat{\pi}^{k}_{\mbox{-}t}: S \times V \rightarrow [0, 1]$ denotes the predictive probability $Pr(v|s)$ of  selecting action $v$ in state $s$ for the other agent $\mbox{-}t$ and is derived using (\ref{eq:m2}) by solving its nested MDPs at levels $0,\ldots,k-1$;
$\phi \in (0, 1)$ is a discount factor; and $b_0 \in \Delta(S)$ is a prior belief over the states of environment.
\vspace{1.5mm}

\noindent
{\bf I-POMDP Lite Planning.} Similar to solving POMDP (except for a few modifications), the optimal value function of I-POMDP Lite for our agent $t$ satisfies the below Bellman equation:
\begin{eqnarray}
V_{n+1}(b) = \displaystyle \max_{u} \Big(R(b, u) + \phi\sum_{v,o}Pr(v,o|b,u)\ V_{n}(b')\Big)
\label{eq:m3b}
\end{eqnarray}
where our agent $t$'s expected immediate payoff is
\begin{equation}
R(b,u) = \displaystyle \sum_{s,v}R(s,u,v)\ Pr(v|s)\ b(s)
\label{eq:m4}
\end{equation}
and the belief update is given as
\begin{equation*}
b'(s') = \beta\ Z(s',u,o)\sum_{s}T(s,u,v,s')\ Pr(v|s)\ b(s)\ .
\label{eq:m4b}
\end{equation*}
Note that (\ref{eq:m4}) yields an intuitive interpretation: The uncertainty over the state of the environment can be factored out of the prediction of the other agent $\mbox{-}t$'s strategy by assuming that agent $\mbox{-}t$ can fully observe the environment. Consequently, solving I-POMDP Lite (\ref{eq:m3b}) involves choosing the policy that maximizes the expected total reward with respect to the prediction of agent $\mbox{-}t$'s mixed strategy using nested MDP. Like POMDP, the optimal value function $V_{n}(b)$ of I-POMDP Lite can be approximated arbitrarily closely (for infinite horizon) by a piecewise-linear and convex function that takes the form of a set $V_n$\footnote{With slight abuse of notation, the value function is also used to denote the set of corresponding $\alpha$ vectors.} of $\alpha$ vectors:
\begin{equation}
V_{n}(b) = \max_{\alpha \in V_{n}} (\alpha\cdot b)\ . 
\label{eq:m7}
\end{equation}
Solving I-POMDP Lite therefore involves computing the corresponding set of $\alpha$ vectors that can be achieved inductively: given a finite set $V_n$ of $\alpha$ vectors, we can plug (\ref{eq:m7}) into (\ref{eq:m3b}) to derive $V_{n+1}$ (see Theorem~\ref{thm:2} in Section~\ref{sect:thr}). Similar to POMDP, the number of $\alpha$ vectors grows exponentially with the time horizon: $|V_{n+1}| = |U||V_n|^{|V||O|}$. To avoid this exponential blow-up, I-POMDP Lite inherits essential properties from POMDP (Section~\ref{sect:thr}) that make it amenable to be solved by existing sampling-based algorithm such as PBVI \cite{Pineau2003} used here. The idea is to sample a finite set $B$ of reachable beliefs (from $b_0$) to approximately represent the belief simplex, thus avoiding the need to generate the full belief reachability tree to compute the optimal policy. This alleviates the curse of history pertaining to the use of interactive beliefs (Section~\ref{sect:intro}). Then, it suffices to maintain a single $\alpha$ vector for each belief point $b \in B$ that maximizes $V_n(b)$. Consequently, each backup step can be performed in polynomial time: $O(|U||V||O||B|^2|S|)$, as sketched below:
\vspace{1.5mm}\\
\noindent{\bf BACKUP}$(V_n, B)$\\
1. $\Gamma^{u,*} \leftarrow \alpha^{u,*}(s) = \sum_v R(s, u, v)\ Pr(v|s)$ \\
2. $\Gamma^{u,v,o} \leftarrow \forall \alpha'_{i} \in V_{n} \ \alpha^{u,v,o}_{i}(s) = \phi\ Pr(v|s)\sum_{s'}Z(s',u,o)\ T(s,u,v,s')\ \alpha'_{i}(s')$\\
3. $\Gamma^{u}_{b} \leftarrow \Gamma^{u,*} + \sum_{v,o} \argmax_{\alpha \in \Gamma^{u,v,o}}(\alpha\cdot b)$ \\
4. Return $V_{n + 1} \leftarrow \forall b\in B \ \argmax_{\Gamma^{u}_{b}, \forall u\in U}(\Gamma^{u}_{b}\cdot b)$
\vspace{1.5mm}

\noindent
{\bf Time Complexity.} Given the set $B$ of sampled beliefs, the cost of solving I-POMDP Lite is divided into two parts: (a) The cost of predicting the mixed strategy of the other agent using nested MDP (\ref{eq:m2}) is $\mathcal{O}\hspace{-0.8mm}\left(kh|S|^2\right)$ (Section~\ref{sect:ip}); (b) To determine the cost of approximately solving I-POMDP Lite with respect to this predicted mixed strategy, since each backup step incurs $\mathcal{O}\hspace{-0.8mm}\left(|U||V||O||B|^2|S|\right)$ time, solving I-POMDP Lite for $h$ steps incurs $\mathcal{O}\hspace{-0.8mm}\left(h|U||V||O||B|^2|S|\right)$ time. By considering $|U|$, $|V|$, and $|O|$ as constants, the cost of solving I-POMDP Lite can be simplified to $\mathcal{O}\hspace{-0.8mm}\left(h|S||B|^2\right)$. Thus, the time complexity of solving I-POMDP Lite is $\mathcal{O}\hspace{-0.8mm}\left(h|S|(k|S| + |B|^2)\right)$, which is much less computationally demanding than the exponential cost of I-PF and I-PBVI (Section~\ref{sect:intro}).
\section{Theoretical Analysis}
\label{sect:thr}
In this section, we prove that I-POMDP Lite inherits convergence, piecewise-linear, and convex properties of POMDP that make it amenable to be solved by existing sampling-based algorithms. More importantly, we show that the performance loss incurred by I-POMDP Lite is linearly bounded by the error of prediction of the other agent's strategy. This result also holds for that of nested MDP policy because I-POMDP Lite reduces to nested MDP under full observability.
\begin{theorem}[Convergence]
Let $V_{\infty}$ be the value function of I-POMDP Lite for infinite time horizon. Then, it is contracting/converging: $\|V_{\infty} - V_{n+1}\|_{\infty} \leq \phi \|V_{\infty} - V_n\|_{\infty}$.
\label{thm:1}
\end{theorem}
\begin{theorem}[Piecewise Linearity and Convexity] 
The optimal value function $V_n$ can be represented as a finite set of $\alpha$ vectors: $V_n(b) = \max_{\alpha \in V_n} (\alpha \cdot b)$.
\label{thm:2}
\end{theorem}
We can prove by induction that the number of $\alpha$ vectors grows exponentially with the length of planning horizon; this explains why deriving the exact I-POMDP Lite policy is intractable in practice.
\begin{definition} 
Let $\pi_{\mbox{-}t}^{\ast}$ be the true strategy of the other agent $\mbox{-}t$ such that $\pi_{\mbox{-}t}^{\ast}(s,v)$ denotes the true probability $Pr^{\ast}(v|s)$ of selecting action $v\in V$ in state $s\in S$ for agent $\mbox{-}t$. Then, the prediction error is $\epsilon_{p} \triangleq\max_{v,s}|Pr^{*}(v|s) - Pr(v|s)|$.
\end{definition}
\begin{definition} 
Let $R_{\max} \triangleq \max_{s,u,v}R(s,u,v)$ be the maximum value of our agent $t$'s payoffs.
\end{definition}
\begin{theorem}[Policy Loss] 
\label{thm:3}
The performance loss $\delta_n$ incurred by executing I-POMDP Lite policy, induced w.r.t the predicted strategy 
$\widehat{\pi}^k_{\mbox{-}t}$ of the other agent $\mbox{-}t$ using nested MDP (as compared to its true strategy $\pi^{\ast}_{\mbox{-}t}$), after $n$ backup steps
is linearly bounded by the prediction error $\epsilon_p:$
\begin{equation}
\delta_n \leq 2\displaystyle\epsilon_p |V| R_{\max}\left[\phi^{n-1} + \frac{1}{1 - \phi}\left(1 + \frac{3\phi|O|}{1 - \phi}\right)\right] \ .\nonumber
\end{equation}
\end{theorem}
The above result implies that, by increasing the accuracy of the prediction of the other agent's strategy, the performance of the I-POMDP Lite policy can be proportionally improved. This gives a very strong motivation to seek better and more reliable techniques, other than our proposed nested MDP framework, for intention prediction. The formal proofs of the above theorems are provided in Appendix D.
\section{Experiments and Discussion}
\label{sect:exp2}
This section first evaluates the empirical performance of nested MDP in a practical multi-agent task called \emph{Intersection Navigation for Autonomous Vehicles} (INAV) (Section~\ref{sect:inav}), which involves a traffic scenario with multiple cars coming from different directions (North, East, South, West) into an intersection and safely crossing it with minimum delay. Our goal is to implement an intelligent autonomous vehicle (\textbf{AV}) that cooperates well with human-driven vehicles (\textbf{HV}) to quickly and safely clear an intersection, in the absence of communication. Then, the performance of I-POMDP Lite is evaluated empirically in a series of partially observable stochastic games (POSGs) (Section~\ref{sect:exp_ipomdp}). All experiments are run on a Linux server with two $2.2$GHz Quad-Core Xeon E$5520$ processors and $24$GB RAM.
\subsection{Nested MDP Evaluations}
\label{sect:inav}
\begin{figure}
\centering
\includegraphics[width=80mm]{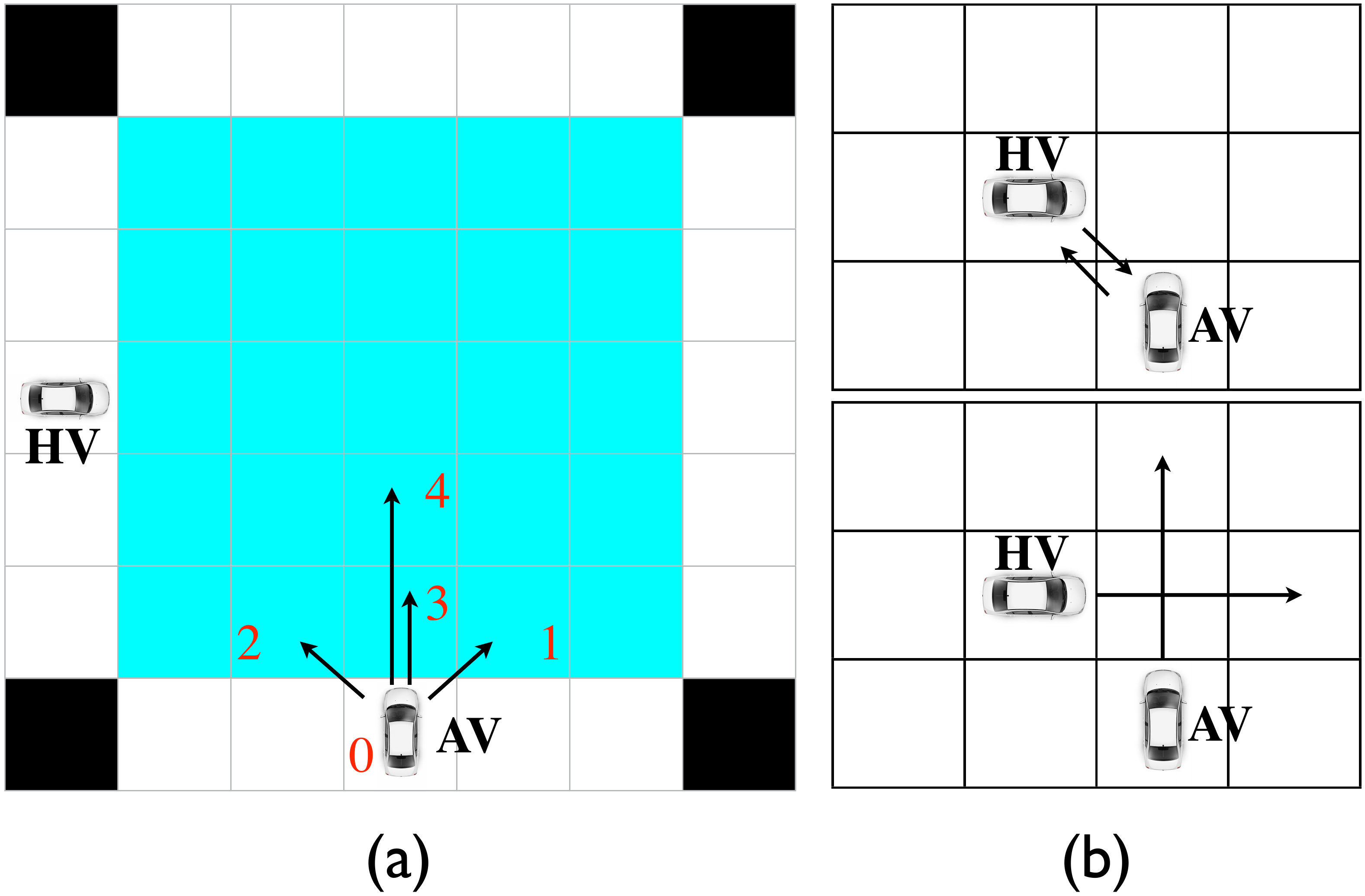}\vspace{-1mm}
\caption{Intersection Navigation: (a) the road intersection modeled as a $7 \times 7$ grid (the black areas are not passable); and (b) accidents caused by cars crossing trajectories.}\vspace{-3mm}
\label{fig:inav}
\end{figure}
In this task, the road intersection is modeled as a $7 \times 7$ grid, as shown in Fig.~\ref{fig:inav}a. The autonomous car (\textbf{AV}) starts at the bottom row of the grid and travels North while a human-driven car (\textbf{HV}) starts at the leftmost column and travels to the East. Each car has five actions: slow down ($0$), forward right ($1$), forward left ($2$), forward ($3$) and fast forward ($4$). Furthermore, it is assumed that `slow down' has speed level $0$, `forward left', `forward', and `forward right' have speed level $1$ while `fast forward' has speed level $2$. The difference in speed levels of two consecutive actions should be at most $1$. In general, the car is penalized by the delay cost $D>0$ for each executed action. But, if the joint actions of both cars lead to an accident by crossing trajectories or entering the same cell (Fig.~\ref{fig:inav}b), they are penalized by the accident cost $C>0$. The goal is to help the autonomous car to safely clear the intersection as fast as possible. So, a smaller value of $D/C$ is desired as it implies a more rational behavior in our agent.

\begin{wrapfigure}{r}{58mm}
\centering
\vspace{-5mm}
\includegraphics[width=0.4\textwidth]{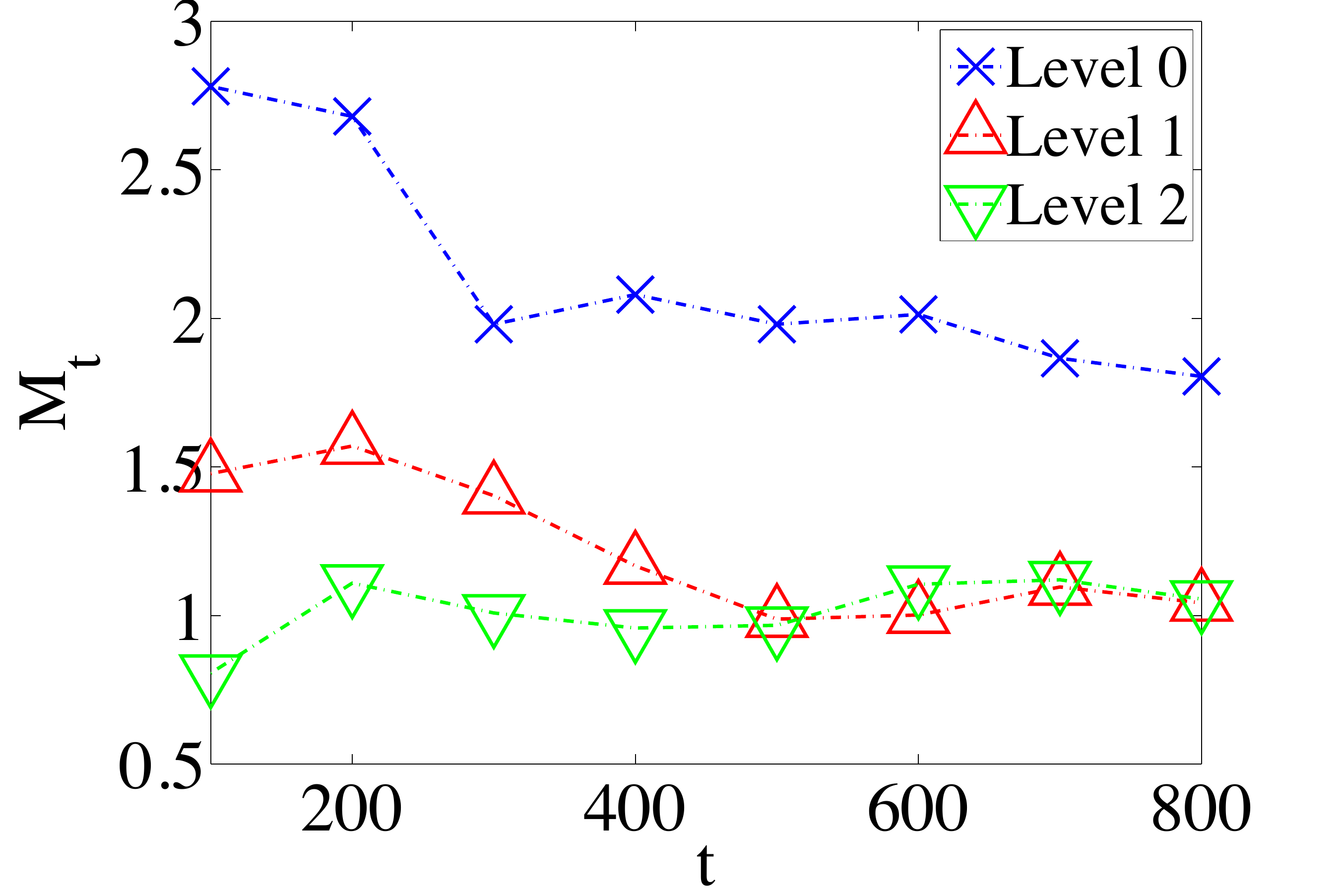}\vspace{-2mm}
\caption{Performance comparison between nested MDPs at reasoning levels 0, 1, and 2.}\vspace{-2mm}
\label{fig:result}
\end{wrapfigure}

The above scenario is modeled using nested MDP, which requires more than $18000$ states. Each state comprises the cells occupied by the cars and their current speed levels. The discount factor $\phi$ is set to $0.99$. The delay and accident costs are hard-coded as $D=1$ and $C=100$. Nested MDP is used to predict the mixed strategy of the human driver and our car's optimal policy is computed with respect to this predicted mixed strategy. For evaluation, our car is run through $800$ intersection episodes. The human-driven car is scripted with the following rational behavior: the human-driven car probabilistically estimates how likely a particular action will lead to an accident in the next time step, assuming that our car selects actions uniformly. It then forms a distribution over all actions such that most of the probability mass concentrates on actions that least likely lead to an accident. Its next action is selected by sampling from this distribution.

We compare the performance of nested MDPs at reasoning levels $k = 0, 1, 2$. When $k = 0$, it is equivalent to the traditional MDP policy that treats the other car as environmental noise. During execution, we maintain a running average $T_t$ (over the first $t$ episodes) of the number of actions taken to clear an intersection and the number $I_t$ of intersections experiencing accidents. The average ratio of the empirical delay is defined as $R_{t}^d = (T_t - T_{\min})/T_{\min} = T_t /T_{\min} - 1$ with $T_{\min} = 3$ (i.e., minimum delay required to clear the intersection). The empirical accident rate is defined as $R_{t}^c = I_t / t$. The average incurred cost is therefore $M_t = CR_{t}^c + DR_{t}^d$. A smaller $M_t$ implies better policy performance. 

Fig.~\ref{fig:result} shows the results of the performance of the evaluated policies.
It can be observed that the $M_t$ curves of nested MDPs at reasoning levels $1$ and $2$ lie below that of MDP policy (i.e., reasoning level $0$). So, nested MDP outperforms MDP. This is expected since our rationality assumption holds: nested MDP's prediction is closer to the human driver's true intention and is thus more informative than the uniformly-distributed human driver's strategy assumed by MDP. Thus, we conclude that nested MDP is effective when the other agent's behavior conforms to our definition of rationality.

\subsection{I-POMDP Lite Evaluations}
\label{sect:exp_ipomdp}
Specifically, we compare the performance of I-POMDP Lite vs. I-POMDP (at reasoning level $k = 1$) players under adversarial environments modeled as zero-sum POSGs. These players are tasked to compete against nested MDP and I-POMDP opponents at reasoning level $k = 0$ (i.e., respectively, MDP and POMDP opponents) whose strategies exactly fit the structural assumptions of I-POMDP Lite (Section~\ref{sect:iap}) and I-POMDP (at $k=1$), respectively. The I-POMDP player is implemented using I-PBVI which is reported to be the best approximate I-POMDP technique \cite{Doshi2008}. Each competition consists of $40$ stages; the reward/penalty is discounted by $0.95$ after each stage. The performance of each player, against its opponent, is measured by averaging its total rewards over $1000$ competitions. Our test environment is larger than the benchmark problems in \cite{Doshi2008}: There are $10$ states, $3$ actions, and $8$ observations for each player. In particular, we let each of the first $6$ states be associated with a unique observation with high probability. For the remaining $4$ states, every disjoint pair of states is associated with a unique observation with high probability. Hence, the sensing capabilities of I-POMDP Lite and I-POMDP players are significantly weaker than that of the MDP opponent with full observability.

Table~\ref{tab:exp} shows the results of I-POMDP and I-POMDP Lite players' performance with varying horizon lengths.
The observations are as follows: (a) Against a POMDP opponent whose strategy completely favors I-POMDP, both players win by a fair margin and I-POMDP Lite  outperforms I-POMDP; (b) against a MDP opponent, I-POMDP suffers a huge loss (i.e., $-37.88$) as its structural assumption of a POMDP opponent is violated, while I-POMDP Lite wins significantly (i.e., $24.67$); and (c) the planning times of I-POMDP Lite and I-POMDP appear to, respectively, grow linearly and exponentially in the horizon length. %
\begin{table}[h!]
\caption{I-POMDP's and I-POMDP Lite's performance against POMDP and MDP opponents with varying horizon lengths $h$ ($|S| = 10, |A| = 3, |O| = 8$). `$\ast$' denotes that the program ran out of memory after $10$ hours.}
\center
\begin{tabular}{lcc!{\vrule width 0.8pt}cc}\noalign{\hrule height 0.8pt}
& POMDP & MDP & Time (s) & $|IS'|$ \\ \hline 
I-POMDP (h = 2) & $13.33$$\pm$$1.75$& $-37.88$$\pm$$1.74$ & $177.35$& $66110$\\ 
I-POMDP (h = 3)& $\ast$ & $\ast$ & $\ast$& $1587010$\\ \hline
I-POMDP Lite (h = 1)& $15.22$$\pm$$1.81$& $15.18$$\pm$$1.41$&$0.02$& N.A. \\
I-POMDP Lite (h = 3)& $17.40$$\pm$$1.71$& $24.23$$\pm$$1.54$& $0.45$ & N.A. \\
I-POMDP Lite (h = 8)& $17.42$$\pm$$1.70$& $24.66$$\pm$$1.54$& $17.11$ & N.A. \\
I-POMDP Lite (h = 10)& $17.43$$\pm$$1.70$& $24.67$$\pm$$1.55$& $24.38$& N.A. \\ \noalign{\hrule height 0.8pt}
\end{tabular}
\vspace{-2mm}
\label{tab:exp}
\end{table}

\begin{figure}
\vspace{-2mm}
\includegraphics[width=140mm]{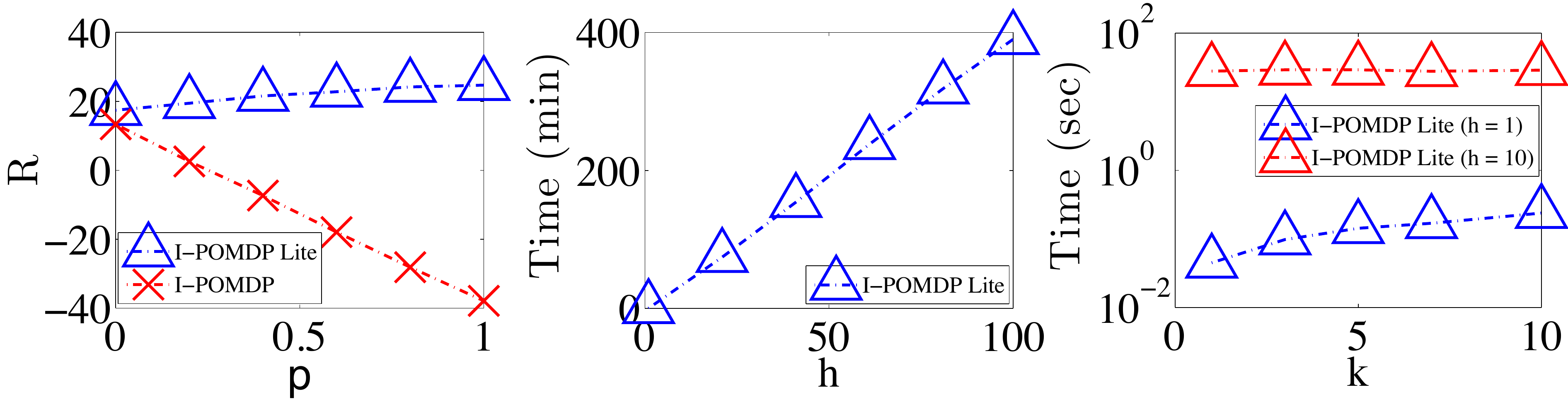}\vspace{-4mm}
\caption{Graphs of (a) performance $R$ of I-POMDP Lite and I-POMDP players against hybrid opponents ($|S| = 10, |A| = 3, |O| = 8$); (b) I-POMDP Lite's planning time vs. horizon length $h$ in a large POSG ($|S| = 100, |A| = 3, |O| = 20$); and (c) I-POMDP Lite's planning time vs. reasoning level $k$ for $h = 1 \text{ and } 10$ ($|S| = 10, |A| = 3, |O| = 10$).}
\label{fig:graphs}
\end{figure}

I-POMDP's exponential blow-up in planning time is expected because its bounded interactive state space $IS'$ increases exponentially in the horizon length (i.e., curse of dimensionality), as shown in Table~\ref{tab:exp}. Such a scalability issue is especially critical to large-scale problems. To demonstrate this, Fig.~\ref{fig:graphs}b shows the planning time of I-POMDP Lite growing linearly in the horizon length for a large zero-sum POSG with $100$ states, $3$ actions, and $20$ observations for each player; it takes about $6$ and $1/2$ hours to plan for $100$-step look-ahead. In contrast, I-POMDP fails to even compute its $2$-step look-ahead policy within $12$ hours.%

It may seem surprising that I-POMDP Lite outperforms I-POMDP even when tested against a POMDP opponent whose strategy completely favors I-POMDP. This can be explained by the following reasons: (a) I-POMDP's exponential blow-up in planning time forbids it from planning beyond $3$ look-ahead steps, thus degrading its planning performance; 
(b) as shown in Section~\ref{sect:iap}, the cost of solving I-POMDP Lite is only polynomial in the horizon length and reasoning depth, thus allowing our player to plan with a much longer look-ahead (Fig.~\ref{fig:graphs}b) and achieve substantially better planning performance;
and (c) with reasonably accurate observation and transition models, Theorem~\ref{thm:0} indicates that the strategy of the MDP opponent (i.e., nested MDP at reasoning level $0$) is likely to approximate that of the true POMDP opponent closely, thus reducing the degree of violation of I-POMDP Lite's structural assumption of a nested MDP opponent.
Such an assumption also seems to make our I-POMDP Lite player overestimate the sensing capability of an unforeseen POMDP opponent and consequently achieve a robust performance against it.
On the other hand, the poor performance of I-POMDP against a MDP opponent is expected because I-POMDP's structural assumption of a POMDP opponent is likely to cause its player to underestimate an unforeseen opponent with superior sensing capability (e.g., MDP) and therefore perform badly against it. In contrast, I-POMDP Lite performs significantly better due to its structural assumption of a nested MDP opponent at level $0$ which matches the true MDP opponent exactly.

Interestingly, it can be empirically shown that when both players' observations are made more informative than those used in the previous experiment, the performance advantage of I-POMDP Lite over I-POMDP, when tested against  a POMDP opponent, increases. To demonstrate this, we modify the previous zero-sum POSG to involve $10$ observations (instead of $8$) such that every state (instead of a disjoint pair of states) is associated with a unique observation with high probability (i.e., $\geq 0.8$); the rest of the probability mass is then uniformly distributed among the other observations. Hence, the sensing capabilities of I-POMDP Lite and I-POMDP players in this experiment are much better than those used in the previous experiment and hence closer to that of the MDP opponent with full observability. Table~\ref{tab:exp2} summarizes the results of I-POMDP Lite's and I-POMDP's performance when tested against the POMDP and MDP opponents in the environment described above.
\begin{table}
\caption{I-POMDP's and I-POMDP Lite's performance against POMDP and MDP opponents with varying horizon lengths $h$ ($|S| = 10, |A| = 3, |O| = 10$). `$\ast$' denotes that the program ran out of memory after $10$ hours.}
\center
\begin{tabular}{lcc!{\vrule width 0.8pt}cc}\noalign{\hrule height 0.8pt}
& POMDP & MDP & Time (s) & $|IS'|$ \\ \hline 
I-POMDP (h = 2) &$5.70$$\pm$$1.67$&$-9.62$$\pm$$1.50$&$815.28$&$102410$\\
I-POMDP (h = 3) & $\ast$&$\ast$&$\ast$&$3073110$\\ \hline 
I-POMDP Lite (h = 1)&$11.18$$\pm$$1.75$&$20.25$$\pm$$1.53$&$0.03$&N.A.\\ 
I-POMDP Lite (h = 3)&$14.89$$\pm$$1.79$&$27.49$$\pm$$1.53$&$0.95$&N.A.\\ 
I-POMDP Lite (h = 8)&$14.99$$\pm$$1.79$&$26.91$$\pm$$1.55$&$24.10$&N.A.\\ 
I-POMDP Lite (h = 10)&$15.01$$\pm$$1.79$&$26.91$$\pm$$1.55$&$33.74$&N.A.\\ \noalign{\hrule height 0.8pt}
\end{tabular}
\vspace{-2mm}
\label{tab:exp2}
\end{table}

To further understand how the I-POMDP Lite and I-POMDP players perform when the sensing capability of an unforeseen opponent varies, we set up another adversarial scenario in which both players pit against a hybrid opponent: At each stage, with probability $p$, the opponent knows the exact state of the game (i.e., its belief is set to be peaked at this known state) and then follows the MDP policy; otherwise, it follows the POMDP policy. So, a higher value of $p$ implies better sensing capability of the opponent. The environment settings are the same as those used in the first experiment, that is, $10$ states, $8$ observations and $3$ actions for each player (Table~\ref{tab:exp}). Fig.~\ref{fig:graphs}a shows the results of how the performance, denoted $R$, of I-POMDP Lite and I-POMDP players vary with $p$: I-POMDP's performance decreases rapidly as $p$ increases (i.e., opponent's strategy violates I-POMDP's structural assumption more), thus increasing the performance advantage of I-POMDP Lite over I-POMDP. This demonstrates I-POMDP Lite's robust performance when tested against unforeseen opponents whose sensing capabilities violate its structural assumption.\\
 
To summarize the above observations, (a) in different partially observable environments where the agents have reasonably accurate observation and transition models, I-POMDP Lite significantly outperforms I-POMDP (Tables~\ref{tab:exp}, \ref{tab:exp2} and Fig.~\ref{fig:graphs}a); and (b) interestingly, it can be observed from Fig.~\ref{fig:graphs}a that when the sensing capability of the unforeseen opponent improves, the performance advantage of I-POMDP Lite over I-POMDP increases. 
These results consistently demonstrate I-POMDP Lite's robust performance against unforeseen opponents with varying sensing capabilities. 
In contrast, I-POMDP only performs well against opponents whose strategies completely favor it, but its performance is not as good as that of I-POMDP Lite due to its limited horizon length caused by the extensive computational cost of modeling the opponent. Unlike I-POMDP's exponential blow-up in horizon length $h$ and reasoning depth $k$ (Section~\ref{sect:intro}), I-POMDP Lite's processing cost grows linearly in both $h$ (Fig.~\ref{fig:graphs}b) and $k$ (Fig.~\ref{fig:graphs}c). When $h = 10$, it can be observed from Fig.~\ref{fig:graphs}c that I-POMDP Lite's overall processing cost does not change significantly with increasing $k$ because the cost $\mathcal{O}\hspace{-0.8mm}\left(kh|S|^2\right)$ of predicting the other agent's strategy with respect to $k$ is dominated by the cost $\mathcal{O}\hspace{-0.8mm}\left(h|S||B|^2\right)$ of solving I-POMDP Lite for large $h$ (Section~\ref{sect:iap}).

\section{Conclusion}
\label{sect:conclude}
This paper proposes the novel nested MDP and I-POMDP Lite frameworks which incorporate the cognitive hierarchy model of games \cite{Camerer2004} for intention prediction into the normative decision-theoretic POMDP paradigm to address some practical limitations of existing planning frameworks for self-interested MAS such as computational impracticality \cite{Doshi2008} and restrictive equilibrium theory of agents' behavior \cite{Hu1998}. We have theoretically guaranteed that the performance losses incurred by our I-POMDP Lite policies are linearly bounded by the error of intention prediction. We have empirically demonstrated that I-POMDP Lite performs significantly better than the state-of-the-art planning algorithms in partially observable stochastic games. Unlike I-POMDP, I-POMDP Lite's performance is very robust against unforeseen opponents whose sensing capabilities violate the structural assumption (i.e., of a nested MDP opponent) that it has exploited to achieve significant computational gain.
In terms of computational efficiency and robustness in planning performance, I-POMDP Lite is thus more practical for use in larger-scale problems.

\bibliographystyle{abbrv}
\bibliography{ijcai2013}

\ifthenelse{\value{sol}=1}{
\clearpage

\appendix

\section{Proof Sketch of Theorem 1}
\label{sketch}

{\bf Theorem 1.} Let $\widehat{Q}_{\mbox{-}t}^{n}(s,v)\triangleq |U|^{-1}\sum_{u\in U}Q_{\mbox{-}t}^{0,n}(s,v,u)$ and $\widehat{Q}_{\mbox{-}t}^{n}(b,v)$ denote $n$-step-to-go values of selecting action $v\in V$ in state $s\in S$ and belief $b$, respectively, for the other agent $\mbox{-}t$ using nested MDP and I-POMDP at reasoning level $0$ (i.e., MDP and POMDP). 
If $b(s) \geq 1 - \epsilon$ and\vspace{-2mm} 
$$\displaystyle\forall \left(s, v, u\right) \exists \left(s', o\right) Pr(s' | s, v, u) \geq 1 - \frac{\epsilon}{2}\ \wedge\ Pr(o|s,v) \geq 1 - \frac{\epsilon}{2}\vspace{-2mm}$$ 
for some  $\epsilon\geq 0$, then\vspace{-1.3mm}
\begin{equation}
\left|\widehat{Q}_{\mbox{-}t}^{n}(s,v) - \widehat{Q}_{\mbox{-}t}^{n}(b,v)\right| \leq \epsilon\ \mathcal{O}\hspace{-0.8mm}\left(\frac{R_{\max} - R_{\min}}{1 - \phi}\right)\vspace{-0.3mm}\label{eq:0}
\end{equation}
where $R_{\max}$ and $R_{\min}$ denote agent $\mbox{-}t$'s maximum and minimum immediate payoffs, respectively.\vspace{1.5mm}

{\bf Proof:} Assuming that \eqref{eq:0} holds with $n$ (it is trivial to verify that it holds with $n = 0$), we need to prove that it also holds with $n + 1$. Define $L_{\mbox{-}t}^{n+1}(s,v)$ as the optimal expected utility of the other agent if it knows that the first state is $s$ and executes $v$ in the first step. From the second step, it only observes the environmental state partially (i.e., same as the POMDP agent). Then,
\begin{eqnarray*}
\left|\widehat{Q}_{\mbox{-}t}^{n+1}(s,v) - \widehat{Q}_{\mbox{-}t}^{n+1}(b,v)\right| &\leq& \left|\widehat{Q}_{\mbox{-}t}^{n+1}(s,v) - L_{\mbox{-}t}^{n+1}(s,v)\right|\\ 
&+& \left|L_{\mbox{-}t}^{n+1}(s,v) - \widehat{Q}_{\mbox{-}t}^{n+1}(b,v)\right|
\end{eqnarray*}

\noindent
{\bf I.} Note that $\widehat{Q}_{\mbox{-}t}^{n+1}(b,v) = \sum_{s'}\alpha_{v}^{*}(s')b(s')$ due to piecewise linearity and convexity and $L_{\mbox{-}t}^{n+1}(s,v) = \alpha_{v}^{*}(s)$, by definition. Assuming that $L_{\mbox{-}t}^{n+1}(s,v) \geq \widehat{Q}_{\mbox{-}t}^{n+1}(b,v)$, the second term in the right-hand side of the above inequality is linearly bounded w.r.t $\epsilon$, as shown below:
\begin{eqnarray*}
L_{\mbox{-}t}^{n+1}(s,v) - \widehat{Q}_{\mbox{-}t}^{n+1}(b,v) &\leq& (\alpha_{v}^{*}(s) - \min_{s' \ne s}\alpha_{v}^{*}(s'))(1 - b(s))\\
&\leq& \epsilon (\alpha_{v}^{*}(s) - \min_{s' \ne s}\alpha_{v}^{*}(s'))\vspace{-1mm}
\end{eqnarray*}
Similarly, if $L_{\mbox{-}t}^{n+1}(s, v) \leq \widehat{Q}_{\mbox{-}t}^{n+1}(b,v)$, we can instead show that $\widehat{Q}_{\mbox{-}t}^{n+1}(b,v) - L_{\mbox{-}t}^{n+1}(s,v) \leq \epsilon(\max_{s' \ne s}\alpha_{v}^{*}(s') - \alpha_{v}^{*}(s))$. Thus, it follows that 
\begin{eqnarray}
\left|L_{\mbox{-}t}^{n+1}(s,v) - \widehat{Q}_{\mbox{-}t}^{n+1}(b,v)\right| &\leq& \epsilon\left(\max_{s'}\alpha_{v}^{*}(s') - \min_{s'}\alpha_{v}^{*}(s')\right) \nonumber \\
&\leq& \epsilon\ \mathcal{O}\hspace{-0.8mm}\left(\frac{R_{\max} - R_{\min}}{1 - \phi}\right)
\label{eq:b1}
\end{eqnarray}

\noindent {\bf II.} From our definitions of $\widehat{Q}_{\mbox{-}t}^{n+1}(s, v)$ and $L_{\mbox{-}t}^{n+1}(s,v)$, it directly follows that\vspace{-2mm}
\begin{equation*}
\widehat{Q}_{\mbox{-}t}^{n+1}(s, v) = \sum_u \frac{1}{|U|}\left(R_{\mbox{-}t}(s, v, u) + \phi\sum_{s'}T_{s}^{u,v}(s')\widehat{V}_{\mbox{-}t}^n(s')\right)\vspace{-1mm}                                                                                                                                                                                                                                                                                                                                                                                                                                                                                                                                                                                                                                                                                                                                                                                                                                                                                                                                                                                                                                                                                                                                                                                                                                                                                                                                                                
\end{equation*}
where $\widehat{V}_{\mbox{-}t}^n(s') \triangleq \max_{v'} \widehat{Q}_{\mbox{-}t}^{n}(s', v')$, $R_{\mbox{-}t}(s,v,u)$ is the other agent's immediate payoff, and $T_s^{u,v}(s') \triangleq Pr(s'|s,v,u)$. Similarly, $L_{\mbox{-}t}^{n+1}(s, v)$ can be expressed as\vspace{-1mm}
\begin{equation*}
L_{\mbox{-}t}^{n+1}(s, v) = \sum_u \hspace{-1mm}\frac{1}{|U|}\hspace{-1mm}\left(\hspace{-1mm} R_{\mbox{-}t}(s, v,u) + \phi\sum_{s',o}T_{s}^{u,v}(s')Z_{s'}^{v}(o)\widehat{V}_{\mbox{-}t}^n(b_{u}^{v,o})\hspace{-1mm}\right)\vspace{-1mm}
\end{equation*}
where $\widehat{V}_{\mbox{-}t}^n(b_{u}^{v,o}) \triangleq \max_{v'}\widehat{Q}_{\mbox{-}t}^{n}(b_{u}^{v,o}, v')$, $Z_{s'}^{v}(o) \triangleq Pr(o|s',v)$, and $\displaystyle b_{u}^{v,o}(s') = \frac{Z_{s'}^{v}(o)T_{s}^{u,v}(s')}{\sum_{s^{''}}Z_{s^{''}}^{v}(o)T_{s}^{u,v}(s^{''})}$\footnote{By definition of $L_{\mbox{-}t}^{n + 1}(s,v)$, the previous state $s$ is known exactly. So, we do not need to take the average w.r.t $b(s)$.}. Then,\vspace{-1mm}
\begin{equation}
\left|\widehat{Q}_{\mbox{-}t}^{n+1}(s,v) - L_{\mbox{-}t}^{n+1}(s,v)\right| \leq \phi\sum_u \frac{1}{|U|}F(u) \label{eq:b2}\vspace{-1mm}
\end{equation}
where $F(u) \triangleq \sum_{s'}T_{s}^{u,v}(s')\left|\widehat{V}_{\mbox{-}t}^n(s') - \sum_{o}Z_{s'}^{v}(o)\widehat{V}_{\mbox{-}t}^n(b_{u}^{v,o})\right|$.
Let $s^*$ be the state at which $\displaystyle T_s^{u,v}(s^*) \geq 1 - \frac{\epsilon}{2}$, we have\vspace{-2mm}
\begin{eqnarray}
F(u) &\leq& \sum_{s' \ne s^*} T_{s}^{u,v}(s')\left|\widehat{V}_{\mbox{-}t}^n(s') - \sum_{o}Z_{s'}^{v}(o)\widehat{V}_{\mbox{-}t}^n(b_{u}^{v,o})\right| \nonumber \\
&+& T_{s}^{u,v}(s^*)\left|\widehat{V}_{\mbox{-}t}^n(s^*) - \sum_{o}Z_{s^*}^{v}(o)\widehat{V}_{\mbox{-}t}^n(b_{u}^{v,o})\right| \label{eq:b3} \\
&\leq& \mathcal{O}\hspace{-0.8mm}\left(\frac{R_{\max} - R_{\min}}{1 - \phi}\right)\sum_{s' \ne s^*}T_{s}^{u,v}(s') \nonumber\\
&+& T_{s}^{u,v}(s^*)\left|\widehat{V}_{\mbox{-}t}^n(s^*) - \sum_{o}Z_{s^*}^{v}(o)\widehat{V}_{\mbox{-}t}^n(b_{u}^{v,o})\right| \label{eq:b4}
\end{eqnarray}
Since $\displaystyle T_s^{u,v}(s^*) \geq 1 - \frac{\epsilon}{2}$ and $\sum_{s' \ne s^*}T_{s}^{u,v}(s') = 1 - T_s^{u,v}(s^*)$, it follows directly that the first term in the right-hand side of \eqref{eq:b4} is bounded by $\displaystyle \frac{\epsilon}{2}\mathcal{O}\hspace{-0.8mm}\left((R_{\max} - R_{\min})/({1 - \phi})\right)$. Then, let $G(s^*, u) \triangleq T_{s}^{u,v}(s^*)\left|\widehat{V}_{\mbox{-}t}^n(s^*) - \sum_{o}Z_{s^*}^{v}(o)\widehat{V}_{\mbox{-}t}^n(b_{u}^{v,o})\right|$ and let $o^*$ be the value at which $\displaystyle Z_{s^*}^{v}(o^*) \geq 1 - \frac{\epsilon}{2}$. 
\begin{eqnarray}
G(s^*, u) &\leq& \left|\widehat{V}_{\mbox{-}t}^n(s^*) - \widehat{V}_{\mbox{-}t}^n(b_{u}^{v,o^*})\right| \nonumber \\
&+& \left|\widehat{V}_{\mbox{-}t}^n(b_{u}^{v,o^*}) - \sum_o Z_{s^*}^{v}(o)\widehat{V}_{\mbox{-}t}^n(b_{u}^{v,o})\right| \label{eq:b7}
\end{eqnarray}
Since the first term on the right-hand side of \eqref{eq:b7} is trivially bounded by $\max_{v'}\left|\widehat{Q}_{\mbox{-}t}^n(s^*, v') - \widehat{Q}_{\mbox{-}t}^n(b_{u}^{v,o^*}, v')\right|$ and the posterior belief $\displaystyle b_{u}^{v,o^*}(s^*) \geq T_s^{u,v}(s^*)Z_{s^*}^{v}(o^*) \geq \left(1 - \frac{\epsilon}{2}\right)^2 \geq 1 - \epsilon$, we can bound it by $\displaystyle \epsilon\mathcal{O}\hspace{-0.8mm}\left((R_{\max} - R_{\min})/({1 - \phi})\right)$ using our inductive assumption. Also, using the same argument in part {\bf I}, the second term is bounded by $\displaystyle \frac{\epsilon}{2}\mathcal{O}\hspace{-0.8mm}\left(\frac{R_{\max} - R_{\min}}{1 - \phi}\right)$. This implies $G(s^*, u)$ and hence $F(u)$ are bounded by $\displaystyle \epsilon\mathcal{O}\hspace{-0.8mm}\left(\frac{R_{\max} - R_{\min}}{1 - \phi}\right)$. Plugging $F(u)$ into \eqref{eq:b2}, we can prove that $\displaystyle \left|\widehat{Q}_{\mbox{-}t}^{n+1}(s,v) - L_{\mbox{-}t}^{n+1}(s,v)\right| \leq \epsilon\mathcal{O}\hspace{-0.8mm}\left(\frac{R_{\max} - R_{\min}}{1 - \phi}\right)$.\vspace{1mm}

\noindent {\bf III.} Putting the results in parts {\bf I} and {\bf II} together, it follows that \eqref{eq:0} holds for $n + 1$ as well.

\section{Empirical Evaluations for Nested MDP}
This section evaluates the performance of nested MDP in a two-player, zero-sum Markov game modeled after soccer. In particular, the soccer game introduced in [Littman, 1994] is played on a $4 \times 5$ grid, as shown in Fig.~\ref{fig:soccer1}. 
\begin{wrapfigure}{r}{3.5cm}
\vspace{-3mm}
\hspace{-2.5mm}
\includegraphics[height=0.1\textheight]{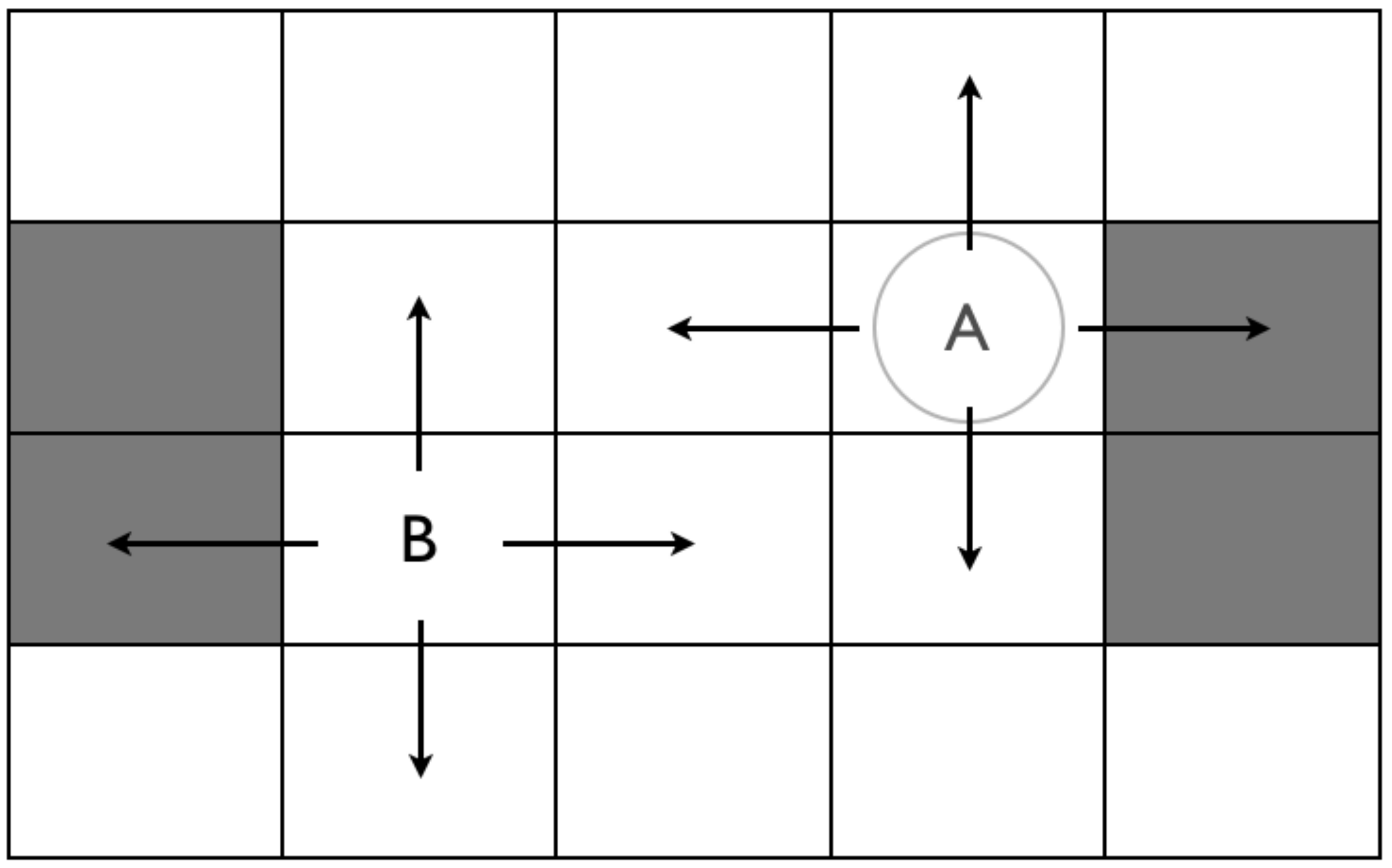}\vspace{-2mm}
\caption{Soccer game.}\vspace{-4mm}
\label{fig:soccer1}
\end{wrapfigure}
At any moment, the two players A and B occupy different squares and have five actions: top, down, left, right, and stand. Only one of them has the ball. Once they have selected their actions, the moves are executed in random order. When the player with the ball moves into its assigned goal (left and right shaded squares for A and B, respectively), it gets one point and its opponent loses a point and vice versa. The game is then reset to the initial configuration shown in Fig.~\ref{fig:soccer1} with the ball possession going to either one at random. When a player steps into the square occupied by its opponent, the ball possession goes to its opponent and the move does not take place. The discount factor is set to $0.9$, which makes scoring sooner better than scoring later. 

The performance of nested MDP player at level $k=1$ is compared against that of the state-of-the-art Nash-Q [Hu and Wellman, 1998] and the MDP planning players. These players are tasked to compete against opponents employing Nash-Q and MDP planning, a random policy selecting actions uniformly, and a hand-built policy specifying simple scoring and blocking rules:

\noindent
{\bf Scoring.} This set of tactics applies when the agent possesses the ball. At any step, if the agent is not on the right track (i.e., its current row goes straight to a goal), it will consider switching to the closest track while avoiding its opponent. Otherwise, it will try to reach the goal as fast as possible without losing the ball to the defending opponent.

Specifically, when the agent is on the right track, it will just keep moving towards the goal if the opponent is not blocking. Otherwise, it will only move forward with a certain probability which is set to be proportional to its distance from the opponent. If the agent decides not to move forward, it will consider staying still with a small probability to maintain the ball. If not, it either moves left or right with equal probability to keep away from the opponent.

When the agent is not on the right track, it simply moves to the closest track if the defending opponent is left behind or too far ahead (i.e., more than two steps ahead). Otherwise, if the opponent is close up ahead, it will consider either moving forward (with the hope of getting past its opponent) or switching to the right track with certain probabilities. Again, we make the agent's probability of moving forward proportional to its distance from the opponent.\vspace{1mm}

\noindent
{\bf Blocking.} When the agent loses the ball, the following tactics apply: The agent will chase after the opponent (who is rushing towards its corresponding goal) if it is left behind. Otherwise, it either switches to the same track (i.e., row) as the opponent or stays still if their tracks are currently the same. That is, the agent always tries to stand in the way of the opponent, which appears to be a very effective defending strategy [Littman, 1994].\vspace{1mm}

Each competition between a player and its opponent consists of a number of games, each of which ends immediately after either the first goal or being declared as a draw with $0.1$ probability at every step. But, each competition only ends after $10000$ games that do not end as a draw. 

Table~\ref{table:compare1} shows the results of the performance of the tested players. The observations are as follows:

\begin{wraptable}{r}{90mm}
\vspace{-5mm}
\begin{scriptsize}
\begin{tabular}{|l|c|c|c|}\hline
\backslashbox{Opponents}{Players}&\textbf{Nash-Q}&\textbf{Nested MDP}&\textbf{MDP}\\ \hline
\textbf{Nash-Q}&$4970\ (173051)$&$4955\ (140867)$&$2393\ (74545)$\\ \hline
\textbf{MDP}&$7602\ (74458)$&$7558\ (53946)$&$4983\ (37277)$\\ \hline
\textbf{Random}&$9771\ (128388)$&$9936\ (96865)$&$9770\ (77992)$\\ \hline
\textbf{Hand-built}&$5575\ (130843)$&$6069\ (99698)$&$2880\ (56854)$\\ \hline
\end{tabular}
\vspace{-3mm}
\caption{Comparison of number of goals scored in a competition by different players: the number of moves taken to complete $10000$ goals are specified in brackets.\vspace{-3mm}}
\label{table:compare1}
\end{scriptsize}
\end{wraptable}
(a) Against a Nash-Q opponent, the nested MDP player's performance is comparable to that of the Nash-Q player. In theory, the best strategy against a Nash-Q opponent should be a corresponding one in the same Nash equilibrium. This implies the nested MDP player achieves a near-optimal policy against the Nash-Q opponent. 
Intuitively, although the nested MDP player does not expect the opponent to play Nash strategy, it can roughly predict which of its opponent's moves would potentially lead to its loss and how likely these moves will be executed. Exploiting this knowledge, it plans a policy that rationally trades off between defense and offense, thereby achieving near-optimal performance against the Nash-Q opponent.

(b) Against a MDP opponent, nested MDP and Nash-Q players' performance are superior to that of the MDP player who wrongly assumes the random action selection by the opponent, thereby leading to an offensive policy with weak defense. Notably, while nested MDP and Nash-Q players' performance are comparable in this case, the nested MDP player completed the competition significantly faster (i.e., fewer moves taken) than the Nash-Q player.

(c) Against a random-policy opponent, Nash-Q and MDP players' performance are almost similar while the nested MDP player performs significantly better. Upon close examination, we observe that, in such a two-player zero-sum game, the Nash-Q strategy coincides with the maximin strategy. So, its corresponding policy is too risk-averse and fails to fully exploit the weaknesses of the random-policy opponent, thereby losing some chances of winning. In contrast, the MDP player performs well because its offensive strategy is generally effective against the weak random-policy opponent. Lastly, since the nested MDP player always has a better balance between defense and offense, its performance is the best among the players. These behaviors can be observed from the number of moves the players took to finish the competition against the same opponent.

(d) Against a hand-built policy opponent, Nash-Q and nested MDP players win more than half the time while the MDP player performs badly due to its ignorance of the opponent's intention, as shown in previous experiments. In particular, the nested MDP player significantly outperforms the Nash-Q player: the nested MDP player, while planning, predicts the opponent's intention by recursively reasoning in its place, thereby recognizing its opponent's critical moves that can potentially lead to its loss. Exploiting this knowledge, the nested MDP player usually has a better trade-off between offense and defense than the Nash-Q player who only focuses on the worst-case situations, thereby losing chances to score if its opponent does not make moves that lead to such situations.

Essentially, these observations show that nested MDP (1) significantly outperforms MDP player (against all opponents) and Nash-Q player (against Random and Hand-built opponents) and (2) performs comparably to Nash-Q player (against Nash-Q and MDP opponents). This suggests that nested MDP is a better planning framework under fully observable settings as compared against the previous ones such as Nash-Q and MDP planning.

\section{Absolute Continuity Condition of Interactive Beliefs for I-PBVI}
\label{b}
In this section, we provide a more detailed discussion on the \emph{Absolute Continuity Condition} (ACC) of interactive beliefs [Doshi and Perez, 2008], which is crucial to the mathematical soundness and hence the feasibility of using I-PBVI. Indeed, we have ensured that this condition is met in all of our experiments so that I-PBVI is not put at a disadvantage against I-POMDP Lite. Intuitively, I-PBVI aims to make the planning of I-POMDP tractable by constraining the infinite interactive state space $IS = S \times \Delta(S)$ to a finite space $IS' = S \times \mbox{Reach}(B, h)$ bounded with respect to the length $h$ of planning horizon. Since $\mbox{Reach}(B,h)$ includes all beliefs that can be reached from $B$ by following a certain action-observation history of length $h$, if the true initial belief of the other agent is included in $B$, then all its reachable beliefs within $h$ steps of interaction are included in $\mbox{Reach}(B,h)$ as well. This essentially means that $IS'$ includes all interactive states of $IS$ that can be assigned a non-zero probability within $h$ steps of interaction and hence guarantees the correctness of the interactive belief update on $IS'$.

On the other hand, if $B$ does not include the other agent's true initial belief, it is possible that the other agent actually reaches an interactive state not included in $\mbox{Reach}(B,h)$ within $h$ steps of interaction. Consequently, this causes the loss of probability mass while updating the interactive belief: if the action-observation history assigns a non-zero probability mass to an interactive state not included in $IS'$, then this mass will be lost because the interactive belief update step only considers those included in $IS'$. As the interaction proceeds, the probability mass will disappear gradually until the whole probability mass, which should sum to $1$, is completely lost (i.e., the actual action-observation history assigns zero probability to all interactive states included in $IS'$). At this point, the agent will become indifferent between the actions and thus perform terribly. We have also verified this by trying to exclude the other agent's true initial belief from $B$: after a few steps, our agent's interactive belief disappears and its performance starts to degrade (i.e., getting negative rewards).

So, the ACC condition simply requires that the other agent's true initial belief should be included in $B$. However, the specification of $B$ is often complicated: if $B$ is made too large just to increase the chance of including the other agent's true initial belief, the planning process may become computationally intractable since $|IS'| = |S| \times |\mbox{Reach}(B,h)| = \mathcal{O}\hspace{-0.8mm}\left(|S||B||U|^h|V|^h|O|^h\right)$. Otherwise, there is a need for highly accurate and informative prior knowledge that can satisfy the ACC condition with a smaller-sized $B$. While such detailed prior knowledge of the other agent is assumed in all our experiments as well as the benchmark problems reported in [Doshi and Perez, 2008], one should be cautioned that it cannot be easily accessed nor processed in real-world situations (e.g., involving humans).
This is another obstacle that limits I-PBVI's practical use.

\section{Proofs}
\label{c}

We hereby present our formal proofs for the theorems stated in Section~\ref{sect:thr}. To increase the readability of this section, we structure our proof in three separate sections:
\squishlisttwo
\item \textbf{Definitions} (Section~\ref{c.1}) -- we introduce the conventional definitions used in this analysis, some of which are previously stated in Section~\ref{sect:thr}.
\item \textbf{Intermediate Results} (Section~\ref{c.2}) -- we state and prove several intermediate results, which are necessary to prove our main theorems.
\item \textbf{Main Theorems} (Section~\ref{c.3}) -- finally, we formally derive the main results stated in Section~\ref{sect:thr}.
\squishend
\subsection{Definitions}
\label{c.1}
\noindent \textbf{Definition 1}. Let $R_{\max} \triangleq \max_{s,u,v}R(s,u,v)$ be the maximum value of our agent's payoffs and let $\pi_{\mbox{-}t}^{\ast}$ be the true mixed strategy of the other agent $\mbox{-}t$ such that $\pi_{\mbox{-}t}^{\ast}(s,v)$ denotes the true probability $Pr^{\ast}(v|s)$ of agent $\mbox{-}t$ selecting action $v\in V$ in state $s\in S$. Then, the prediction error is\vspace{-2mm}
$$\displaystyle\epsilon_{p} \triangleq\max_{v,s}|Pr^{*}(v|s) - Pr(v|s)| \ .$$\vspace{-4mm}

\noindent \textbf{Definition 2}. Given the true mixed strategy $\pi^{*}_{\mbox{-}t}(s,v) \triangleq Pr^{*}(v|s)$ of the agent $\mbox{-}t$, let us denote $Pr^{*}(v,o|b,u)$, $B^{*}(b,u,v,o)$ and $V^{*}_{n}(b)$ as the belief-state observation model, belief-state update function and optimal value function over time horizon $n$ computed with respect to $Pr^{*}(v|s)$.\\

\noindent \textbf{Definition 3}. Let $V_n(b)$ denotes the optimal value function computed with respect to the prediction $Pr(v|s)$. The difference between $V_n(b)$ and the true optimal value function $V^{*}_n(b)$ is $\delta_{n} \triangleq \max_b|V^{*}_n(b) - V_n(b)|$.\\

\noindent \textbf{Definition 4}. Let $Q^{*}_n(b, u)$ and $Q_n(b, u)$ denote the corresponding Q-functions of $V^{*}_{n}(b)$ and $V_n(b)$:
\begin{eqnarray}
V^{*}(b) &=& \max_u Q^{*}_n(b,u) \nonumber \\
V_n(b) &=& \max_u Q_n(b, u) \ . \nonumber
\end{eqnarray}
\noindent Consequently, we define the optimal policies $\pi^*$ and $\pi$ induced from $V^{*}_n$ and $V_n$, respectively, as:
\begin{eqnarray}
\pi^*(b) &=& \argmax_u Q^{*}_n(b, u) \nonumber \\
\pi(b) &=& \argmax_u Q_n(b, u) \ . \nonumber
\end{eqnarray}

\noindent \textbf{Definition 5}. Let $J^{*}_n(b)$ and $J_n(b)$ denote the expected total rewards if, starting from belief $b$, we follow the optimal policy $\pi^*$ and the induced policy $\pi$ for $n$ steps, respectively. Thus, we have:
\begin{eqnarray}
J^{*}_n(b) &=& \sum_{s, v}b(s)Pr^{*}(v|s)R(s, \pi^*(b), v) \nonumber \\
                 &+& \phi\sum_{v, o}Pr^{*}(v,o|b,\pi^*(b))J^{*}_{n-1}(B^{*}(b, \pi^*(b), v, o)) \nonumber \\
J_n(b) &=& \sum_{s, v}b(s)Pr^{*}(v|s)R(s, \pi(b), v) \nonumber \\ 
           &+& \phi\sum_{v, o}Pr^{*}(v,o|b,\pi(b))J_{n-1}(B^{*}(b, \pi(b), v, o)) \ . \nonumber
\end{eqnarray}
\noindent Consequently, it can be verified by the definition of the value function $V^{*}_n$ that $J^{*}_n \equiv V^{*}_n$.

\subsection{Immediate Results}
\label{c.2}
\noindent \textbf{Lemma 1}. Given a prediction $\pi_{\mbox{-}t}(s,v) \triangleq Pr(v|s)$ of the other agent $\mbox{-}t$'s mixed strategy, the difference between the optimal value function $V_n$ and the true optimal value function $V^{*}_n$ of our agent $t$, with respect to agent $\mbox{-}t$'s true mixed strategy $\pi^{*}_{\mbox{-}t}(s,v) \triangleq Pr^{*}(v|s)$, is bounded by:
\begin{eqnarray}
\delta_n \leq \max_b\max_u |Q^{*}_{n}(b, u) - Q_n(b, u)| \ .
\end{eqnarray}
\noindent \textbf{Proof}: Let us denote $u' = \argmax_u Q_n(b, u)$ and $u^{*} = \argmax_u Q^{*}_{n}(b, u)$. Assume that $V^{*}_{n}(b) \leq V_n(b)$,  we have
\begin{eqnarray}
|V^{*}_{n}(b) - V_n(b)| &=& V_n(b) - V^{*}_{n}(b) = Q_n(b, u') - \max_{u}Q^{*}_{n}(b, u) \nonumber \\
&\leq& Q_n(b, u') - Q^{*}(b, u') = |Q_n(b, u') - Q^{*}_{n}(b, u')| \label{eq:1}
\end{eqnarray}
\noindent Similarly, if $V_n(b) \leq V^{*}_{n}(b)$ we also have
\begin{eqnarray}
|V^{*}_{n}(b) - V_n(b)| &\leq& |Q^{*}_n(b, u^{*}) - Q_n(b, u^{*})| \label{eq:2}
\end{eqnarray}
\noindent Therefore, from inequalities \eqref{eq:1} and \eqref{eq:2}, we have
\begin{eqnarray}
|V^{*}_{n}(b) - V_n(b)| &\leq& \max \left(|Q_n(b, u') - Q^{*}_{n}(b, u')|, |Q^{*}_n(b, u^{*}) - Q_n(b, u^{*})| \right) \nonumber \\
                                                      &\leq&  \max_u |Q^{*}_{n}(b, u) - Q_n(b, u)| \label{eq:3}
\end{eqnarray}
\noindent From \eqref{eq:3} and Definition 3, we have $\delta_n \leq \max_b\max_u |Q^{*}_{n}(b, u) - Q_n(b, u)| \text{ }\Box$\\

\noindent \textbf{Lemma 2}. Suppose that at current belief $b$, our agent (i.e., agent $t$) executes action $u$. We defined the expected immediate payoffs with respect to our predictive distribution $\pi_{\mbox{-}t}(s,v) \triangleq Pr(v|s)$ and the true mixed strategy $\pi^{*}_{\mbox{-}t}(s,v) \triangleq Pr^{*}(v|s)$ of agent $\mbox{-}t$ as
\begin{eqnarray}
R^{*}(b, u) &=& \sum_s b(s) \sum_v R(s, u, v)Pr^{*}(v|s) \nonumber \\
R(b, u) &=& \sum_s b(s) \sum_v R(s, u, v) Pr(v|s) \ . \nonumber
\end{eqnarray} 
\noindent The difference between $R^{*}(b, u)$ and $R(b, u)$ is linearly bounded by the prediction error $\epsilon_p$:
\begin{eqnarray}
|R^{*}(b, u) - R(b, u)| \leq \epsilon_{p}|V|R_{\max} \ . \nonumber
\end{eqnarray}
\noindent \textbf{Proof}: We have
\begin{eqnarray}
|R^{*}(b, u) - R(b, u)| &\leq& \sum_s b(s) \sum_v R(s,u,v) |Pr^{*}(v|s) - Pr(v|s)| \nonumber \\
&\leq& \sum_s b(s) \sum_v R(s,u,v)\epsilon_p  \leq \sum_s b(s)|V|R_{\max} \epsilon_p \nonumber \\
&\leq& |V| R_{\max} \epsilon_p \ .
\end{eqnarray}
\noindent $\Box$\\

\noindent \textbf{Lemma 3}. The difference between the belief-state observation models $Pr^{*}(v, o|b, u)$ and $Pr(v, o|b, u)$ with respect to agent $\mbox{-}t$'s true mixed strategy $\pi^{*}_{\mbox{-}t}(s,v) \triangleq Pr^{*}(v|s)$ and our predictive distribution $\pi_{\mbox{-}t}(s,v) \triangleq Pr(v|s)$ is bounded by the prediction error $\epsilon_p$:\vspace{-0mm}
\begin{eqnarray}
|Pr^{*}(v,o|b,u) - Pr(v,o|b,u)| \leq \epsilon_p \ . \nonumber
\end{eqnarray}\vspace{-0mm}

\noindent \textbf{Proof}: We have
\begin{eqnarray}
|Pr^{*}(v,o|b,u) - Pr(v,o|b,u)| &\leq& \sum_{s'}Z(s',u,o)\sum_{s}T(s,u,v,s')b(s)|Pr^{*}(v|s) - Pr(v|s)| \nonumber \\
&\leq& \epsilon_p\sum_{s}b(s)\sum_{s'}Z(s',u,o)T(s,u,v,s') \nonumber \\
&\leq& \epsilon_p\sum_{s}b(s)\sum_{s'}T(s,u,v,s') = \epsilon_p\sum_{s}b(s) = \epsilon_p \ .
\end{eqnarray}
\noindent $\Box$\\

\noindent \textbf{Lemma 4}. For any two beliefs $b$ and $b'$, if $\|b-b'\|_{1} \leq \delta$, we have $|V(b) - V(b')| \leq \frac{R_{\max}}{1 - \phi}\delta \ .$ (Lipschitz condition)

\noindent \textbf{Proof}: For any two beliefs $b$ and $b'$, let $\alpha = \argmax_{\alpha \in V}(\alpha b)$ and $\alpha' = \argmax_{\alpha \in V}(\alpha b')$. Without loss of generality, assuming that $V(b') \leq V(b)$, we have
\begin{eqnarray}
|V(b) - V(b')| &=& \alpha b - \alpha'b' \nonumber \\
&\leq& \alpha b - \alpha b' = \alpha(b - b') \nonumber \\
&=& \sum_{s}\alpha(s)(b(s)-b'(s)) \nonumber \leq \sum_{s}\alpha(s)|(b(s)-b'(s)| \nonumber \\
&\leq& \max_{s}\alpha(s)\sum_{s}|b(s) - b'(s)| \leq \max_{s}\alpha(s)\delta \nonumber \\
&\leq& \frac{R_{\max}}{1 - \phi}\delta \ .
\end{eqnarray}
\noindent The last step follows from the fact that each component $\alpha(s)$ of an $\alpha$ vector is basically an expected total reward (if the initial state is $s$ and the agent follows the optimal policy) which is always bounded by $\frac{R_{max}}{1 - \phi}\text{ }\Box$\\

\noindent \textbf{Lemma 5}. Let us define the unnormalized belief update function $F(b,u,v,o)$ of $B(b,u,v,o)$ as the followings:
\begin{eqnarray}
F(b,u,v,o)(s') &=& Z(s',u,o)\sum_{s}T(s,u,v,s')Pr(v|s)b(s) \nonumber \\
                       &=& B(b,u,v,o)Pr(v,o|b,u) \ . \nonumber
\end{eqnarray}

\noindent The norm-$1$ distance between the unnormalized belief update functions $F^{*}(b,u,v,o)$ and $F(b,u,v,o)$ of $B^{*}(b,u,v,o)$ and $B(b,u,v,o)$, respectively, is at most $\epsilon_p$.\\

\noindent \textbf{Proof}: Let us shortly write $F(b, u, v, o)$ and $F^{*}(b, u, v, o)$ as $f'$ and $f^*$, respectively. We have
\begin{eqnarray}
\|f' - f^{*}\|_{1} &=& \sum_{s'}Z(s',u,o)\sum_{s}T(s,u,v,s')b(s)|Pr(v|s) - Pr^{*}(v|s)| \nonumber \\
&\leq& \epsilon_p\sum_{s'}Z(s',u,o)\sum_{s}T(s,u,v,s')b(s) \nonumber \\
&=& \epsilon_p\sum_{s}b(s)\sum_{s'}Z(s',u,o)T(s,u,v,s') \nonumber \\
&\leq& \epsilon_p\sum_{s}b(s)\sum_{s'}T(s,u,v,s') \nonumber \\
&=& \epsilon_p\sum_{s}b(s) = \epsilon_p \ . 
\end{eqnarray}
\noindent $\Box$\\

\noindent \textbf{Proposition 1}. For all values $b \in B, u \in U, v \in V, o \in O$, we have
\begin{eqnarray}
Pr^{*}(v,o|b,u)\|B^{*}(b,u,v,o) - B(b,u,v,o)\|_{1} \leq 2\epsilon_p \ . \nonumber
\end{eqnarray}
\noindent \textbf{Proof}: Let $b^{*}$ and $b'$ denote $B^{*}(b,u,v,o)$ and $B(b,u,v,o)$, respectively. Also, let $f^{*}$ and $f'$ be unnormalized version of $b^{*}$ and $b'$. Finally, let $A^{*}$ and $A$ denote the belief-state observation probability $Pr^{*}(v,o|b,u)$ and $Pr(v,o|b,u)$. Thus, we have
\begin{eqnarray}
A^{*}\|b^{*} - b'\|_{1} &=& A^{*}\sum_{s'}\left|\frac{f^{*}(s')}{A^{*}} - \frac{f'(s')}{A}\right| \nonumber \\
&=& \frac{1}{A}\sum_{s'}\left| f^{*}(s')A - f'(s')A^{*}\right| \nonumber \\
&=& \frac{1}{A}\sum_{s'}\left| A(f^{*}(s') - f'(s')) + f'(s')(A - A^{*})\right| \nonumber \\
&\leq& \frac{1}{A}\sum_{s'}\left( A|f^{*}(s')-f'(s')| + f'(s')|A - A^{*}|\right) \nonumber \\
&\leq& \frac{1}{A}\sum_{s'}\left( A|f^{*}(s')-f'(s')| + f'(s')\epsilon_p\right) \text{  (Lemma 3)} \nonumber \\
&=& \sum_{s'}|f^{*}(s') - f'(s')| + \epsilon_p\frac{1}{A}\sum_{s'}f'(s') \nonumber \\
&=& \sum_{s'}|f^{*}(s') - f'(s')| + \epsilon_p\frac{1}{A}\sum_{s'}Ab'(s') \text{  (Def. of $f'(.)$)} \nonumber \\
&=& \sum_{s'}|f^{*}(s') - f'(s')| + \epsilon_p \nonumber \\
&\leq& \epsilon_p + \epsilon_p = 2\epsilon_p \text{  (Lemma 5)}
\end{eqnarray} 
\noindent $\Box$\\

\noindent \textbf{Proposition 2}. Given a predictive distribution $\pi_{\mbox{-}t}(s,v) \triangleq Pr(v|s)$ of agent $\mbox{-}t$'s mixed strategy, we have the following inequality:
\begin{eqnarray}
\left| V^{*}(B^{*}(b,u,v,o))Pr^{*}(v,o|b,u) - V^{*}(B(b,u,v,o))Pr(v,o|b,u) \right| \leq 3\epsilon_p\frac{R_{\max}}{1 - \phi} \nonumber
\end{eqnarray}
\noindent \textbf{Proof}: Let $b^{*}$ and $b'$ denote $B^{*}(b,u,v,o)$ and $B(b,u,v,o)$, respectively. Also, let $A^{*}$ and $A$ denote the probability $Pr^{*}(v,o|b,u)$ and $Pr(v,o|b,u)$. Thus, we have:
\begin{eqnarray}
\left|V^{*}(b^{*})A^{*} - V^{*}(b')A\right| &\leq& \left| V^{*}(b^{*})A^{*} - V^{*}(b')A^{*} \right| + \left| V^{*}(b')A^{*} - V^{*}(b')A\right| \nonumber \\
&=& A^{*}\left| V^{*}(b^{*}) - V^{*}(b') \right| + V^{*}(b')\left| A^{*} - A \right| \nonumber \\
&\leq& A^{*}\frac{R_{\max}}{1 - \phi}\|b^{*} - b'\|_{1} + V^{*}(b')\left| A^{*} - A \right| \text{  (Lemma 4)} \nonumber \\
&=& \frac{R_{\max}}{1 - \phi}A^{*}\|b^{*} - b'\|_{1} + V^{*}(b')\left| A^{*} - A \right| \nonumber \\
&\leq& 2\epsilon_p\frac{R_{\max}}{1 - \phi} + V^{*}(b')\left| A^{*} - A \right| \text{  (Proposition 1)} \nonumber \\
&\leq& 2\epsilon_p\frac{R_{\max}}{1 - \phi} + V^{*}(b')\epsilon_p \text{  (Lemma 3)} \nonumber \\
&\leq& 2\epsilon_p\frac{R_{\max}}{1 - \phi} + \epsilon_p\frac{R_{\max}}{1 - \phi} = 3\epsilon_p\frac{R_{\max}}{1 - \phi} \ .
\end{eqnarray}
\noindent The last step follows from the fact that the value function $V^{*}(b')$ is essentially the expected total reward if the agent follows the optimal policy from the initial belief. Thus, it is trivially bounded by $\frac{R_{\max}}{1 - \phi} \text{ }\Box$\\

\noindent \textbf{Proposition 3}. For all values $b \in B, u \in U$, we have: 
\begin{eqnarray}
\left|Q^{*}_{n}(b,u) - Q_{n}(b,u)\right| &\leq& \phi\delta_{n - 1} + \epsilon_p |V| R_{\max}\left( 1 + 3\phi\frac{|O|}{1 - \phi} \right) \ . \nonumber
\end{eqnarray} 

\noindent \textbf{Proof}: Let $b^{*}$ and $b'$ denote $B^{*}(b,u,v,o)$ and $B(b,u,v,o)$, respectively. Further, let us define
\begin{eqnarray}
L_{n}(b,u) &=& R(b,u) + \phi\sum_v\sum_o V^{*}_{n-1}(b')Pr(v,o|b,u) \ . \nonumber
\end{eqnarray}
\noindent First, we prove that $|L_{n}(b, u) - Q_{n}(b,u)| \leq \phi\delta_{n - 1}$ :
\begin{eqnarray}
|L_{n}(b, u) - Q_{n}(b,u)| &\leq& \phi\sum_{v,o} Pr(v,o|b,u)|V^{*}_{n-1}(b') - V_{n-1}(b')| \nonumber \\
&\leq& \phi\sum_{v,o} Pr(v,o|b,u)\delta_{n-1} \text{  (by def. of $\delta_{n}$)} \nonumber \\
&\leq& \phi\delta_{n-1}\sum_{v, o} Pr(v,o|b,u) = \phi\delta_{n-1} \ . \label{eq:4}
\end{eqnarray}
\noindent Second, we prove that $|Q^{*}_{n}(b, u) - L_{n}(b,u)| \leq \epsilon_p |V| R_{\max}\left( 1 + 3\phi\frac{|O|}{1 - \phi} \right)$:
\begin{eqnarray}
|Q^{*}_{n}(b, u) - L_{n}(b,u)| &\leq& |R^{*}(b,u) - R(b,u)| + \nonumber \\
& & \phi\sum_{v,o} \left| V^{*}(b^*)Pr^{*}(v,o|b,u) - V^{*}(b')Pr(v,o|b,u) \right| \nonumber \\
&\leq& \epsilon_p |V| R_{\max} + \phi\sum_{v,o}\left(3\epsilon_p\frac{R_{\max}}{1 - \phi}\right) \text{  (Lemma 2, Proposition 2)} \nonumber \\
&=&  \epsilon_p |V| R_{\max} + 3\epsilon_p|V||O|\frac{\phi R_{\max}}{1 - \phi} \nonumber \\
&=&  \epsilon_p |V| R_{\max}\left( 1 + 3\phi\frac{|O|}{1 - \phi} \right) \ . \label{eq:5}
\end{eqnarray} 
\noindent Finally, we prove the main result of this proposition:
\begin{eqnarray}
\left|Q^{*}_{n}(b,u) - Q_{n}(b,u)\right| &\leq& |Q^{*}_{n}(b, u) - L_{n}(b,u)| + |L_{n}(b, u) - Q_{n}(b,u)| \nonumber \\
&\leq& \phi\delta_{n - 1} + \epsilon_p |V| R_{\max}\left( 1 + 3\phi\frac{|O|}{1 - \phi} \right) \ .
\end{eqnarray}
\noindent The last step follows from inequalities \eqref{eq:4} and \eqref{eq:5} $\Box$

\noindent \textbf{Proposition 4}. Given our agent's prediction $\pi_{\mbox{-}t}(s,v) \triangleq Pr(v|s)$ of its counterpart's mixed strategy (i.e., agent $\mbox{-}t$'s mixed strategy) with prediction error $\epsilon_p$, the incurred error $\delta_{n}$ of the corresponding optimal value function $V_n$, with respect to $Pr(v|s)$, is linearly bounded by $\epsilon_p$:
\begin{eqnarray}
\delta_{n} &\leq& \displaystyle\epsilon_p |V| R_{\max}\left[\phi^{n-1} + \frac{1}{1 - \phi}\left(1 + \frac{3\phi|O|}{1 - \phi}\right)\right] \ . \nonumber
\end{eqnarray}
\noindent \textbf{Proof}: By Lemma 1, we have:
\begin{eqnarray}
\delta_{n} &\leq& \max_b\max_u |Q^{*}_{n}(b,u) - Q_{n}(b,u)| \ . \label{eq:6}
\end{eqnarray}
Also, Proposition 3 shows that:
\begin{eqnarray}
\left|Q^{*}_{n}(b,u) - Q_{n}(b,u)\right| &\leq& \phi\delta_{n - 1} + \epsilon_p |V| R_{\max}\left( 1 + 3\phi\frac{|O|}{1 - \phi} \right) \ . \label{eq:7}
\end{eqnarray}
From \eqref{eq:6} and \eqref{eq:7}, we have:
\begin{eqnarray}
\delta_{n} &\leq& \phi\delta_{n - 1} + \epsilon_p |V| R_{\max}\left( 1 + 3\phi\frac{|O|}{1 - \phi} \right) \nonumber \\
&\leq& \phi^{n - 1}\delta_{1} + \frac{\epsilon_p |V| R_{\max}}{1 - \phi}\left( 1 + 3\phi\frac{|O|}{1 - \phi} \right) \text{  (expanding the recurrence)} \nonumber \\
&\leq& \phi^{n - 1}\epsilon_p |V| R_{\max} + \frac{\epsilon_p |V| R_{\max}}{1 - \phi}\left( 1 + 3\phi\frac{|O|}{1 - \phi} \right) \ . 
\end{eqnarray}
\noindent When $n \rightarrow \infty$, $\phi^{n - 1}\epsilon_p |V| R_{\max} \rightarrow 0$. Thus, with large value of $n$ the incurred error is approximately bounded by $\frac{\epsilon_p |V| R_{\max}}{1 - \phi}\left( 1 + 3\phi\frac{|O|}{1 - \phi} \right)$. Consequently, when $\epsilon_p \rightarrow 0$, $\delta_{n} \rightarrow 0$. In general, this implies that the prediction error is linearly proportional with the incurred error of the corresponding value function $\Box$

\subsection{Main Theorems}
\label{c.3}
\noindent \textbf{Theorem 2.} Let $V_{\infty}$ be the value function for infinite time horizon. Then, we have $\|V_{\infty} - V_{n+1}\|_{\infty} \leq \phi\|V_{\infty} - V_n\|_{\infty}$.\\

\noindent \textbf{Proof}: We have
\begin{eqnarray}
|V_{\infty}(b) - V_{n+1}(b)| &\leq& \max_u |Q_{\infty}(b, u) - Q_{n+1}(b,u)|\\
&\leq& \max_u \phi \sum_{v,o}Pr(v,o|b,u)|V_{\infty}(b') - V_n(b)|\\
&\leq& \max_u \phi \sum_{v,o}Pr(v,o|b,u)\|V_{\infty} - V_n\|_{\infty}\\
&=& \phi \|V_{\infty} - V_n\|_{\infty} \max_u \sum_{v,o}Pr(v,o|b,u) \\
&=& \phi \|V_{\infty} - V_n\|_{\infty}
\end{eqnarray}
\noindent The last equation completes our proof. $\Box$\\

\noindent \textbf{Theorem 3.} The optimal value function $V_n$ of I-POMDP Lite is a piecewise-linear and convex function represented as a finite set of $\alpha$ vectors:
\begin{eqnarray}
V_n(b) = \max_{\alpha \in V_n} (\alpha b) \ . \label{eq:5.4}
\end{eqnarray}
\noindent \textbf{Proof}: For $n = 1$, it can be verified that the set of $\alpha$ vectors $V_n$ are simply the weighted average immediate payoffs:
\begin{eqnarray}
V_1 = \left\{\alpha_u \mid u \in U, \forall s \in S, \alpha_u(s) = \sum_v Pr(v|s)R(s, u, v)\right\} \ . \nonumber
\end{eqnarray}
\noindent Inductively, let us assume that $V_n$ is a piecewise-linear and convex function up to $n = k$. We need to show that $V_{k+1}$ is also a piecewise-linear and convex function, represented as a finite set of $\alpha$ vectors.\\

\noindent To simplify the notation, let us denote $R(s, u) = \sum_v Pr(v|s)R(s, u, v)$ and $b' = B(b,u,v,o)$. Also, let us index the $\alpha$ vectors of $V_k$ with the set of numberings $I = \{1, 2, \ldots, |V_k|\}$. Subsequently, we have
\begin{eqnarray}
V_{k+1}(b) &=& \max_u \left(\sum_{s}b(s)R(s,u) + \phi\sum_{v,o}Pr(v,o|b,u)\max_{i \in I}\sum_{s'}b'(s')\alpha'_{i}(s')\right) \ .\hspace{4mm} \label{eq:5.5}
\end{eqnarray}
\noindent Recall that the belief update step is defined as
\begin{eqnarray}
b'(s') &=& \frac{1}{Pr(v,o|b,u)}Z(s', u, o)\sum_s T(s, u, v, s')Pr(v|s)b(s) \ .\nonumber
\end{eqnarray}
\noindent Let $l'$ be the unnormalized version of $b'$, we have
\begin{eqnarray}
l'(s') &=& Z(s', u, o)\sum_s T(s, u, v, s')Pr(v|s)b(s) \ . \label{eq:5.6}
\end{eqnarray}
\noindent Using equation \eqref{eq:5.6}, we can simplify equation \eqref{eq:5.5} as
\begin{eqnarray}
V_{k+1}(b) &=& \max_u \left(\sum_s b(s)R(s,u) + \phi\sum_{v,o}\max_{i \in I}\sum_{s'}\alpha'_{i}(s')l'(s')\right) \nonumber \\
                   &=& \max_u \left(\sum_s b(s)R(s,u) + \phi\max_{x_{1,1} \in I}\max_{x_{1,2} \in I}\ldots\max_{x_{|V|,|O|} \in I}\sum_{v,o,s'}\alpha'_{x_{v,o}}(s')l'(s')\right)\nonumber \\
                   &=& \max_u\max_{x_{1,1} \in I}\ldots\max_{x_{|V|,|O|} \in I}\left(\sum_s b(s)R(s,u) + \phi\sum_{v,o,s'}\alpha'_{x_{v,o}}(s')l'(s')\right) \ . \label{eq:5.7} 
\end{eqnarray}
\noindent Substitute equation \eqref{eq:5.6} into \eqref{eq:5.7}, we obtain
\begin{eqnarray}
V_{k+1}(b) = \max_u\max_{x_{1,1} \in I}\ldots\max_{x_{|V|,|O|} \in I}\left(\sum_s b(s) \left(R(s,u) + \phi\sum_{v,o,s'}\alpha'_{x_{v,o}}(s')Z(s',u,o)T(s,u,v,s')Pr(v|s)\right)\right).\nonumber
\end{eqnarray}
\noindent The last equation shows that $V_{k+1}(b)$ is a piecewise-linear and convex function. Essentially, it is equivalent to equation \eqref{eq:5.4}. $\Box$\\

\noindent This proof also implies an exponential increase in the number of $\alpha$ vectors after each back-up operation. Intuitively, for each tuple $(u, x_{1,1}, \ldots, x_{|V|, |O|})$, we can compute a new $\alpha$ vector for $V_{k+1}$. Thus, the number of $\alpha$ vectors needed to exactly represent $V_{k+1}$ is equal to the number of those tuples. Since the domain value for each variable $x_{i,j}$ is the set of integers from $I = \left\{1, 2, \ldots, |V_k|\right\}$ and there are $|V||O|$ of those variables, we have $|U||V_k|^{|V||O|}$ of such tuples. Consequently, there will be $|V_{k+1}| = |U||V_k|^{|V||O|}$ $\alpha$ vectors generated to represent $V_{k+1}$. This explains why exact back-up operation with respect to the whole belief simplex is infeasible in practice.\\

\noindent \textbf{Theorem 4}. The performance loss $\delta_n \triangleq \|J^{*}_n - J_n\|_{\infty}$ incurred by executing I-POMDP Lite policy, induced with respect to the predicted mixed strategy $\pi_{\mbox{-}t}(s,v) \triangleq Pr(v|s)$ of agent $\mbox{-}t$ (as compared to its true mixed strategy $\pi^\ast_{\mbox{-}t}(s,v) \triangleq Pr^{*}(v|s)$), after $n$ backup steps
is linearly bounded by the prediction error $\epsilon_p$:
\begin{eqnarray}
\|J^{*}_n - J_n\|_{\infty} &\leq& \displaystyle 2\epsilon_p |V| R_{\max}\left[\phi^{n-1} + \frac{1}{1 - \phi}\left(1 + \frac{3\phi|O|}{1 - \phi}\right)\right] \ . \nonumber 
\end{eqnarray}

\noindent \textbf{Proof}: Let us define $I_n(b)$ as the expected total reward if our agent follows the optimal policy $\pi$ computed with respect to our predictive distribution $Pr(v|s)$, and if the other agent's true mixed strategy is exactly $Pr(v|s)$. We have:
\begin{eqnarray}
I_n(b) = \sum_{s,v}Pr(v|s)R(s, \pi(b), v)b(s) + \phi\sum_{v,o}Pr(v,o|b, \pi(b))I_{n-1}(b') \ . \nonumber
\end{eqnarray}
\noindent with $b' = B(b, \pi(b), v, o)$.\\

\noindent It can be trivially verified that $I_n \equiv V_n$. Now, we have
\begin{eqnarray}
|J^{*}_n(b) - J_n(b)| &\leq& |J^{*}_n(b) - I_n(b)| + |I_n(b) - J_n(b)| \nonumber \\
&=& |V^{*}_n(b) - V_n(b)| + |I_n(b) - J_n(b)| \nonumber \\
&\leq& \epsilon_pC + |I_n(b) - J_n(b)| \text{  (Proposition 4)} \ . \label{eq:8}
\end{eqnarray}
\noindent with $C$ is a constant that represents for:
\begin{eqnarray}
C &=& \phi^{n-1}|V|R_{\max} + \frac{R_{\max}|V|}{1 - \phi}\left(1 + \frac{3\phi|O|}{1 - \phi}\right) \ . \label{eq:9} 
\end{eqnarray}
\noindent For convenience, let us denote from now on that $u = \pi(b)$ and $b^* = B^{*}(b, \pi(b), v, o)$. Thus, we have
\begin{eqnarray}
|I_n(b) - J_n(b)| &\leq& \sum_{s,v}b(s)R(s, u, v)|Pr^{*}(v|s) - Pr(v|s)| + \nonumber \\
& & \phi\sum_{v,o}\left| Pr^{*}(v,o|b, u)J_{n-1}(b^*) - Pr(v,o|b,u)I_{n-1}(b')\right| \nonumber \\
&\leq& \phi\sum_{v,o}\left| Pr^{*}(v,o|b, u)J_{n-1}(b^*) - Pr(v,o|b,u)I_{n-1}(b')\right| + \nonumber \\
& & R_{\max}|V|\epsilon_p \text{  (Lemma 2)} \label{eq:10}
\end{eqnarray}
\noindent Next, we have
\begin{eqnarray}
\left| Pr^{*}(v,o|b, u)J_{n-1}(b^*) - Pr(v,o|b,u)I_{n-1}(b') \right| &\leq& Pr^{*}(v,o|b,u)|J_{n-1}(b^*) - I_{n-1}(b')| + \nonumber \\
& & I_{n-1}(b')|Pr^{*}(v,o|b,u) - Pr(v,o|b,u)| \nonumber \\
&\leq& Pr^{*}(v,o|b,u)|J_{n-1}(b^*) - I_{n-1}(b')| + \nonumber \\
& & I_{n-1}(b')\epsilon_p \text{  (Lemma 3)} \nonumber \\
&\leq& Pr^{*}(v,o|b,u)|J_{n-1}(b^*) - I_{n-1}(b')| + \nonumber \\
& & \frac{R_{\max}}{1 - \phi}\epsilon_p \label{eq:11}
\end{eqnarray}
\noindent The above last step follows because the expected total reward $I_{n-1}(b')$ is always bounded by $ \frac{R_{\max}}{1 - \phi}$. Let us denote $\gamma_n = \max_b |I_n(b) - J_n(b)|$, we have
\begin{eqnarray}
Pr^{*}(v,o|b,u)|J_{n-1}(b^*) - I_{n-1}(b')| &\leq& Pr^{*}(v,o|b,u)\left(|J_{n-1}(b^*) - I_{n-1}(b^*)| + |I_{n-1}(b^*) - I_{n-1}(b')|\right) \nonumber \\
&\leq& Pr^{*}(v,o|b,u)\left(|J_{n-1}(b^*) - I_{n-1}(b^*)| + |V_{n-1}(b^*) - V_{n-1}(b')|\right) \nonumber \\
&\leq& Pr^{*}(v,o|b,u)\left(|J_{n-1}(b^*) - I_{n-1}(b^*)| + \frac{R_{\max}}{1 - \phi}\|b^* - b'\|\right) \text{  (Lemma 4)} \nonumber \\
&\leq& Pr^{*}(v,o|b,u)\left(\gamma_{n-1} + \frac{R_{\max}}{1 - \phi}\|b^* - b'\|\right) \text{  (Def. of $\gamma_n$)} \nonumber \\
&\leq& Pr^{*}(v,o|b,u)\gamma_{n-1} + \frac{R_{\max}}{1 - \phi}Pr^{*}(v,o|b,u)\|b^* - b'\| \nonumber \\
&\leq& Pr^{*}(v,o|b,u)\gamma_{n-1} + \frac{R_{\max}}{1 - \phi}2\epsilon_p \text{  (Proposition 1)} \nonumber \\
&=& Pr^{*}(v,o|b,u)\gamma_{n-1} + 2\epsilon_p\frac{R_{\max}}{1 - \phi} \label{eq:12}
\end{eqnarray}
\noindent Substitute \eqref{eq:12} into \eqref{eq:11}, we have
\begin{eqnarray}
\left| Pr^{*}(v,o|b, u)J_{n-1}(b^*) - Pr(v,o|b,u)I_{n-1}(b') \right| \leq Pr^{*}(v,o|b,u)\gamma_{n-1} + 3\epsilon_p\frac{R_{\max}}{1 - \phi} \ . \label{eq:13}
\end{eqnarray}
\noindent Substitute inequality \eqref{eq:13} into \eqref{eq:10}, we have
\begin{eqnarray}
|I_n(b) - J_n(b)| &\leq& R_{\max}|V|\epsilon_p + \phi\sum_{v,o}\left(Pr^{*}(v,o|b,u)\gamma_{n-1} + 3\epsilon_p\frac{R_{\max}}{1 - \phi}\right) \nonumber \\
&\leq& R_{\max}|V|\epsilon_p + \phi\sum_{v,o}Pr^{*}(v,o|b,u)\gamma_{n-1} + 3\phi|V||O|\frac{R_{\max}}{1 - \phi}\epsilon_p \nonumber \\
&\leq& \phi\gamma_{n-1} + R_{\max}|V|\epsilon_p + 3\phi|V||O|\frac{R_{\max}}{1 - \phi}\epsilon_p \ . \label{eq:14}
\end{eqnarray}
\noindent Since inequality \eqref{eq:14} holds for all $b$, we have the recurrence equation
\begin{eqnarray}
\gamma_n &\leq& \phi\gamma_{n-1} + \epsilon_p|V|R_{\max}\left(1 + \frac{3\phi|O|}{1 - \phi}\right) \nonumber \\
&\leq&  \phi^{n-1}\gamma_1 + \frac{\epsilon_p|V|R_{\max}}{1 - \phi}\left(1 + \frac{3\phi|O|}{1 - \phi}\right) \text{  (expand the recurrence)} \nonumber \\
&\leq& \phi^{n-1}\epsilon_p|V|R_{\max} + \frac{\epsilon_p|V|R_{\max}}{1 - \phi}\left(1 + \frac{3\phi|O|}{1 - \phi}\right) \nonumber \\
&=& \epsilon_p\left(\phi^{n-1}|V|R_{\max} + \frac{R_{\max}|V|}{1 - \phi}\left(1 + \frac{3\phi|O|}{1 - \phi}\right)\right) \nonumber \\
&=& \epsilon_pC \ . \label{eq:15}
\end{eqnarray}
\noindent From \eqref{eq:15}, it is obvious that $|I_n(b) - J_n(b)| \leq \gamma_n \leq \epsilon_pC$. Hence, plugging it into inequality \eqref{eq:8}, we have
\begin{eqnarray}
|J^{*}_n(b) - J_n(b)| \leq \epsilon_pC + |I_n(b) - J_n(b)| \leq 2\epsilon_pC \ . \label{eq:16}
\end{eqnarray}
\noindent Since inequality \eqref{eq:16} holds for all $b$, we have:
\begin{eqnarray}
\|J^{*}_n - J_n\|_{\infty} = \max_b |J^{*}_n(b) - J_n(b)| \leq 2\epsilon_pC \ . \label{eq:17}
\end{eqnarray}
\noindent Finally, we complete the proof by substituting equation \eqref{eq:9} into inequality \eqref{eq:17}:
\begin{eqnarray}
\|J^{*}_n - J_n\|_{\infty} &\leq& 2\left(\phi^{n - 1}\epsilon_p |V| R_{\max} + \frac{\epsilon_p |V| R_{\max}}{1 - \phi}\left(1 + 3\phi\frac{|O|}{1 - \phi} \right)\right) \ . 
\end{eqnarray}
\noindent $\Box$

}{}

\end{document}